\documentclass{article}

\usepackage{microtype}
\usepackage{graphicx}
\usepackage{subcaption}
\usepackage{booktabs} % for professional tables

\usepackage{hyperref}

\usepackage[preprint]{icml2026}

\usepackage{amsmath}
\usepackage{amssymb}
\usepackage{mathtools}
\usepackage{amsthm}

\usepackage[capitalize,noabbrev]{cleveref}

\theoremstyle{plain}

\theoremstyle{definition}

\theoremstyle{remark}

\usepackage[textsize=tiny]{todonotes}

\usepackage{etoolbox}
\usepackage{xspace}
\usepackage{multirow}
\usepackage[most]{tcolorbox}
\newcommand{\newpara}[1]{\vspace{0.0cm}\noindent\textbf{#1}}
\usepackage{enumitem}

\newcommand{\lcbio}{\text{LCB-IO}\xspace}
\newcommand{\dmcvt}{\text{DMC}\xspace}
\newcommand{\crux}{\text{CruxEval}\xspace}
\newcommand{\cruxo}{\text{CruxEval-O}\xspace}
\newcommand{\codelms}{\text{code LLMs}\xspace}

\newcommand{\nlex}{\text{NLEX}\xspace}

\usepackage{tcolorbox}
\usepackage{listings}
\tcbuselibrary{listings, skins, breakable}

\tcbset{
  promptstyle/.style={
    enhanced,
    colback=blue!5!white,
    colframe=blue!50!black,
    coltitle=blue!30!black,
    colbacktitle=blue!10!white,
    fonttitle=\bfseries,
    boxrule=0.8pt,
    arc=6pt,
    top=-1.5mm, bottom=-1.5mm,
    left=1mm, right=1mm,
  }
}

\newtcblisting{stylecode}[1][]{
  promptstyle,
  listing only,
  listing options={
    language=Python,
    basicstyle=\ttfamily\small,
    keywordstyle=\color{blue}\bfseries,
    stringstyle=\color{red},
    commentstyle=\color{green!50!black},
    keepspaces=true,
    breaklines=true,
    showstringspaces=false,
    tabsize=4
  },
  title=Code Prompt,
  #1
}

\newtcblisting{styletext}[1][]{
  promptstyle,
  listing only,
  breakable,
  listing options={
    language={}, % Removes language specific highlighting
    basicstyle=\ttfamily\small,
    keepspaces=true,
    breaklines=true,
    showstringspaces=false
  },
  title=Text Prompt,
  #1
}

\icmltitlerunning{Self-Execution Simulation Improves Coding Models}
\begin{document}

\twocolumn[
  \icmltitle{Self-Execution Simulation Improves Coding Models}



  \icmlsetsymbol{equal}{*}

  \begin{icmlauthorlist}
    \icmlauthor{Gallil Maimon}{comp,huji}
    \icmlauthor{Ori Yoran}{comp}
    \icmlauthor{Felix Kreuk}{comp}
    \icmlauthor{Michael Hassid}{comp,huji}
    \icmlauthor{Gal Cohen}{comp}
    \icmlauthor{Pierre Chambon}{comp,sch}
    \icmlauthor{Yossi Adi}{comp,huji}
  \end{icmlauthorlist}

  \icmlaffiliation{comp}{FAIR team, Meta}
  \icmlaffiliation{huji}{Hebrew University of Jerusalem}
  \icmlaffiliation{sch}{Inria}

  \icmlcorrespondingauthor{Gallil Maimon}{gallil.maimon@mail.huji.ac.il}

  \icmlkeywords{Code LLM, execution feedback, code world model}

  \vskip 0.3in
]



\printAffiliationsAndNotice{}  % no special notice (required even if empty)

\begin{abstract}
A promising research direction in enabling LLMs to generate consistently correct code involves addressing their inability to properly estimate program execution, particularly for code they generate. In this work, we demonstrate that \codelms can be trained to simulate program execution in a step-by-step manner and that this capability can be leveraged to improve competitive programming performance. Our approach combines supervised fine-tuning on natural language execution traces, textual explanations grounded in true execution, with reinforcement learning using verifiable rewards. We introduce two complementary objectives: output prediction given code and inputs, and solving competitive programming tasks with either ground-truth or self-predicted execution feedback. These objectives enable models to perform \emph{self-verification} over multiple candidate solutions, and iterative \emph{self-fixing} by simulating test execution. Across multiple competitive programming benchmarks, our method yields consistent improvements over standard reasoning approaches. We further present ablations and analysis to elucidate the role of execution simulation and its limitations.
\end{abstract}

\section{Introduction} 

Going beyond treating code as a static text block holds great promise in advancing \codelms. This involves jointly modelling program syntax and execution dynamics, similar to how developers reason during debugging and development~\citep{armengol2025cannot, licodeio, thimmaiah2025plsemanticsbench, cwm, beck2026neuraldebugger}. 
Despite its promise, translating execution prediction capabilities into consistent gains on practical programming tasks remains an open challenge. Moreover, \citet{gu2024counterfeit,silver_bullet,kamoi2024can} indicate that current models often fail to faithfully simulate runtime behaviour or to consistently identify and explain errors in code they generate.

Code execution is widely used in various parts of training and inference of \codelms. This includes feedback from code execution~\citep{rlef,11052794} or rich tool-based signals in agentic settings~\citep{xia2025live}. However, executing code at scale for training or inference poses practical challenges, such as environment setup~\cite{bogin2024super}, managing code dependencies~\cite{jimenez2023swe}, handling partial or non-executable code, and sandboxing. In broader settings, program execution can also be computationally expensive and time-consuming; for example, runs of MLE-Bench can take up to $9$ hours~\citep{chan2024mle, zheng2026can}. Predicting execution outcomes could mitigate these challenges by enabling large rollouts and policy optimisation without code execution~\citep{minimax2026m21, kimi}. More broadly, using execution prediction to support reasoning and planning in coding tasks can be viewed as a form of world modelling in the code domain~\cite{ha2018world, ding2025understanding}.

\begin{figure*}[t!]
  \begin{center}
    \centerline{\includegraphics[width=0.95\textwidth]{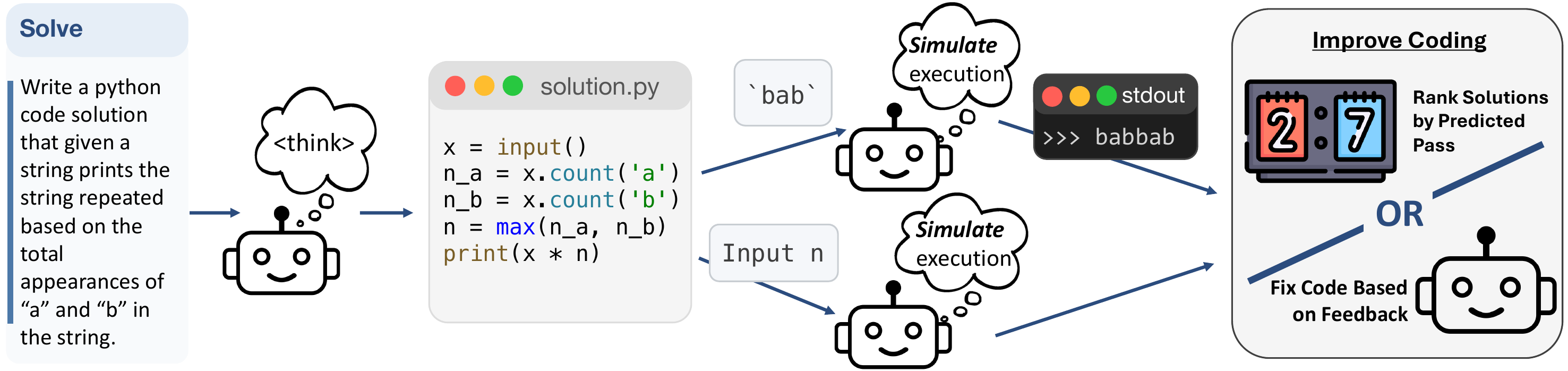}}
    \caption{A conceptual outline of how one can use \emph{self-execution simulation} of a generated code solution (or solutions) on public or generated test cases to improve coding performance. The simulation can be used as feedback to select the best solution from a few candidates (best@k) or to iteratively fix the code as needed (self-RLEF). See Section \ref{sec:solving_with_sim} for details.}
    \label{fig:teaser}
  \end{center}
\end{figure*}

In this work, we take a step in this direction. We show LLMs can learn to simulate program execution step-by-step, including code they generated, and use this capability to improve competitive programming performance. We start by training models on \textit{natural-language execution traces} -- text explanations grounded in real program executions -- and then further refining them using single-turn reinforcement learning for code output prediction. Equipped with this capability, we empirically demonstrate how models can perform \emph{self-verification} over parallel solutions based on simulated execution (best@k). Inspired by \citet{rlef}, we also design a multi-turn reinforcement learning pipeline that enables iterative \emph{self-fixing} through code proposal, execution simulation, and refinement. \cref{fig:teaser} provides a conceptual overview of the proposed methods.

Results suggest the proposed training recipe leads to significant improvements in output prediction on CruxEval~\citep{cruxeval} (up to $43\%$) and competitive programming solutions~\cite{dmc, lcb} (up to $39\%$) relative to the evaluated baseline. This applies to both external and self-generated code solutions. Under the best@k setting, using the model's output prediction to verify its own candidate solutions improves code correctness by up to $5.5\%$ absolute points on competitive programming tasks. In the multi-turn setting, we observe consistent gains across all evaluated configurations. Compared to ground-truth execution, both best@k and multi-turn variants show a relatively small degradation. Finally, we conduct analysis to highlight the strengths and limitations of the proposed approach.

\newpara{Our Contributions:} We provide a training recipe, showing that \codelms are capable of simulating the program execution for both external and self-generated code. With that in mind, we introduce a practical framework for leveraging this behaviour by filtering code solutions based on predicted output (i.e., \textit{self-verification}). Lastly, we present a multi-turn training and inference process to perform iterative \textit{self-fixing} of the model's generated code.
\begin{figure*}[t]
  \begin{center}
    \centerline{\includegraphics[width=0.95\textwidth]{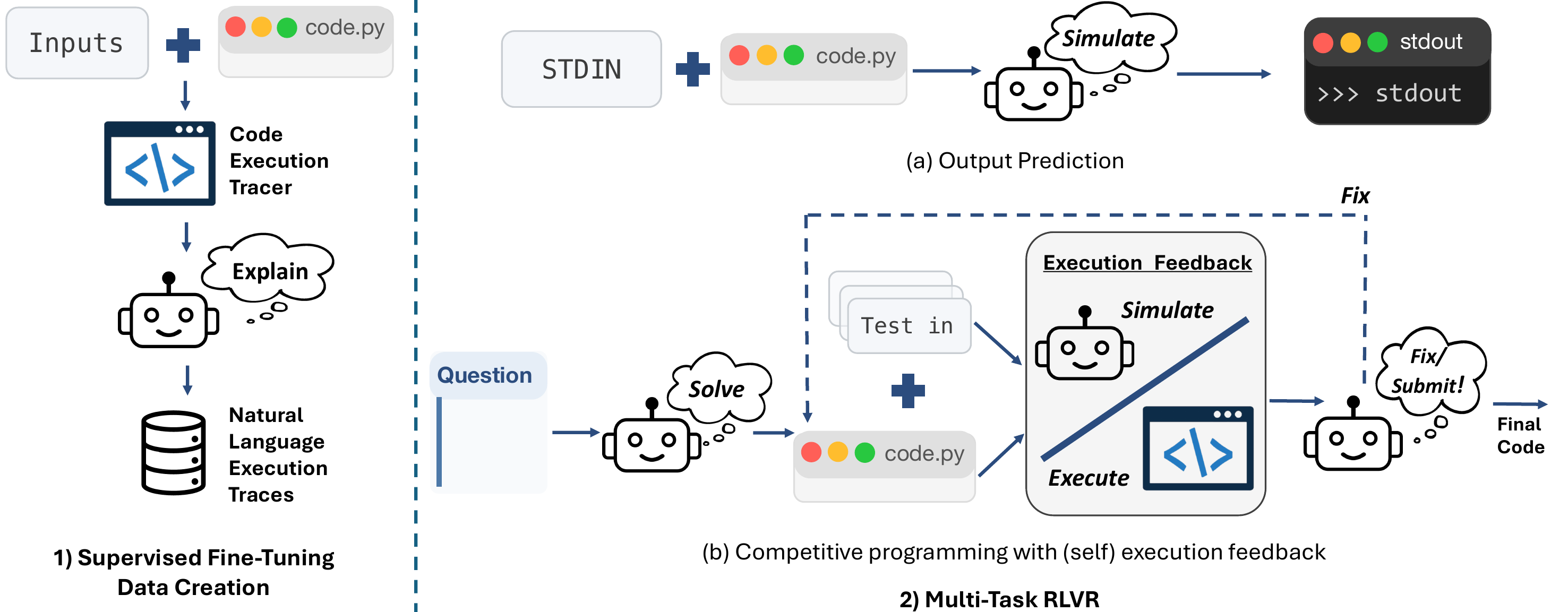}}
    \caption{The two parts of our training pipeline. 1) Supervised fine tuning on natural language execution traces (\nlex), 2) Multi-task reinforcement learning on output prediction and competitive programming (optionally with multi-turn feedback and fixing).}
    \label{fig:approach}
  \end{center}
\end{figure*}

\section{Boosting Execution Simulation}\label{sec:out_pred}
Following \citet{cwm}, we collect executable Python programs with input–output pairs and record their line-by-line execution. Next, we convert these execution traces into natural-language explanations and use the resulting data for supervised fine-tuning. We then further train the model using verifiable rewards on an output prediction task. The next sections describe these post-training steps in detail.

\subsection{Natural Language Execution Tracing (\nlex)} \label{sec:nlex}

We collect Python functions from public repositories and automatically synthesise suitable inputs using a combination of LLM prompting and lightweight fuzzing techniques. In addition, we collect LLM-generated solutions to competitive programming problems from CodeContests~\cite{dmc}, and keep their provided tests as inputs. Although this portion of the data is smaller in scale, it involves substantially more complex programs. We then record execution traces for each program–input pair, capturing intermediate variable states throughout execution. Following~\citet{cwm}, we exclude traces exceeding $10 \text{k}$ events or requiring more than $1 \text{MB}$ of storage. The resulting dataset comprises ${\sim}30 \text{M}$ functions from basic code sources and $35 \text{k}$ from competitive programming problems. For all of the above, we use Llama3-70B-Instruct \cite{llama3}. 

While CWM \cite{cwm} focused primarily on a structured, JSON-like format to describe the step-by-step execution, we wish to focus on natural language description of this data. Relative to the structured format, a free-form variant holds several benefits. First, as based on natural language, it closely matches the reasoning-style data already used by LLMs. It also enables adding semantic context to operations, e.g., explaining an update to an array in the scope of a dynamic programming code. Finally, it naturally abstracts away unnecessary details, such as summarising a long loop that reverses strings character by character.

To this end, we prompt Qwen3-32B-FP8 (without thinking) \cite{yang2025qwen3} to ``translate'' execution traces from raw structured format to a natural language explanation. See Appendix \ref{app:prompts} for the exact prompt. We discard instances where the model's predicted output does not match the ground truth, resulting in ${\sim}80$ M execution descriptions for general Python functions and $115$ k for competitive programming solutions (notice, each traced function may contain several io-pairs). The resulting data is formatted as instruction-following examples and used for model fine-tuning during the SFT phase. In which, the user requests a step-by-step explanation of a program’s execution for a given input, and the assistant provides the translated explanation. Sample instances are provided in Appendix~\ref{app:samples}.

\subsection{Output Prediction Environment}\label{sec:out_pred_rl}
Following standard practice in reasoning models, we post-train our model using Reinforcement Learning with Verifiable Rewards (RLVR). We define an output prediction environment, based on coding tasks, where the model reasons over a given (\texttt{code}, \texttt{stdin}) pair to predict the resulting \texttt{stdout}. We employ a terminal binary reward, scoring $+1$ if the prediction matches the true \texttt{stdout}, and $-1$ otherwise, allowing $1e-5$ tolerance in float comparisons.

The intended downstream use of the output prediction ability is simulating the execution of model generated solutions to competitive programming questions. To that end, we construct the output prediction environment on precisely such data. We collect solutions from strong LLMs to competitive programming questions and use the \texttt{stdin} of the matching public tests. Moreover, the higher difficulty of competitive programming problems makes them particularly well suited for post-training. \cref{fig:approach} depicts the optimisation pipeline. 

\section{Self-Execution For Verification}
\label{sec:solving_with_sim}

Given models with increased ability to simulate code execution, we ask \emph{``How can this ability be used to boost programming abilities?"} Arguably, the simplest and most straightforward approach to leverage such capability is through post-hoc solution filtering. In this approach, candidate solutions are simulated on public or generated tests and retained only if their predicted outputs align with the expected ones. 

For that, we adopt a best-of-$k$ (best@$k$) evaluation setup, where the model independently samples $k$ candidate solutions and selects the final one based on predefined criteria. In our setup, selection is based on the model's execution prediction. In other words, for each candidate, the model simulates its execution on public test cases and checks whether the predicted outputs match the expected ones. The candidate \textit{predicted} to pass the greatest number of public tests is selected for submission. In case of a tie we randomly select a solution among the ones that passed the maximum number of tests. We denote this approach \textit{best@k simulate}. Notice, during inference we do not access any private tests nor ground-truth verification. 

Formally, given a set of solutions $\mathcal{S}$, with public input-output pairs $(in_t, out_t) \in \mathcal{T}$, we use a model to simulate execution, predict the output $\mathcal{M}_{\text{sim}}(s, in_t)$, and select:

\[
\mathrm{Best}( \mathcal{S} ) \coloneqq
\operatorname*{arg\,max}_{s \in \mathcal{S}}
\sum_{(in_t, out_t) \in \mathcal{T}}
\mathbf{1}\!\left[
\mathcal{M}_{\text{sim}}(s, in_t) = out_t
\right].
\]

We use \verb|rank_score_at_k|~\cite{larger} to provide an unbiased accuracy estimate for generating $k$ solutions and selecting the one with the highest score under the proposed heuristic. Specifically, we use $20$ generated solutions per task and $5$ output-prediction attempts per test.

Recall, the primary focus of this work is \emph{self}-simulation. In which, the same LLM is used to both generate candidate solutions and simulate their execution. That said, the same method can also be applied to solutions produced by other models. In~\cref{sec:res}, we present empirical evidence demonstrating the efficacy of this approach in both setups.

\section{Self-Execution For Fixing}
\label{sec:multi-turn}
Another approach to leveraging execution feedback is through multi-turn interaction with a computational environment to perform code fixing. \citet{rlef} demonstrated that exposing LLMs to environmental feedback can enhance programming performance by allowing models to iteratively refine solutions based on information from failed test cases. However, as mentioned above, this may introduce practical challenges such as environment configuration, code dependencies, and non-executable code.

Motivated by this paradigm, we introduce an approach that uses \textit{predicted} execution outputs as feedback instead of actual program execution. Note, unlike the method presented in \cref{sec:solving_with_sim}, that verifies multiple solutions via self-execution, the multi-turn setup refines solutions sequentially based on predicted feedback. Ideally, this approach can leverage richer signals, such as past solutions and execution details. 
While similar world-modelling approaches have been explored in vision, recent work shows limited gains from such signals~\citep{qian2026current}. In contrast, we show that using execution simulation can improve performance.

\begin{algorithm}[t]
\caption{Multi-Turn Self-RLEF Rollout}
\label{alg:rollout}
\begin{algorithmic}[1]
\REQUIRE model $M$, question $q$, public tests $T_{pub} = \{(in_t, out_t)\}$

\STATE $s \leftarrow M(q)$ \COMMENT{Generate an initial code solution}
\STATE $k \leftarrow 1$

\WHILE{$k \leq K_{\max}$}
    \FOR{$(in_t, out_t) \in T_{pub}$}
        \STATE $ \hat{out}_t \leftarrow M(s, in_t)$
        \COMMENT{self execution simulation}
    \ENDFOR

    \STATE $sub\_fix \leftarrow M(q, s, \{(in_t, out_t, \hat{out}_t)\})$
    \COMMENT{Submit current code as correct or provide new solution}

    \IF{$sub\_fix = \texttt{submit}$}
        \STATE \textbf{Return} $s$
    \ELSE
        \STATE $s \leftarrow sub\_fix$
    \ENDIF
    \STATE $k \leftarrow k + 1$
\ENDWHILE
\end{algorithmic}
\end{algorithm}

Specifically, we propose a multi-turn environment with explicit \textit{context switching}, i.e. where each interaction step is represented as an independent single-turn prompt containing only the relevant information (see details in the bullets below). This design enables fine-grained control over information flow. For instance, ensuring that execution simulation is isolated from solution reasoning and from access to the correct outputs. Moreover, it mitigates long-context challenges commonly associated with multi-turn reasoning~\citep{yao2022react}. Finally, this context switching also naturally allows one to extend the number of fix turns at inference as each fix turn is independent. A formal description of the rollout procedure is provided in Algorithm~\ref{alg:rollout}, and an example inference of our model in Appendix \ref{app:srlef_example}. 
In words, the multi-turn setup is designed as follows:

\begin{itemize}[nosep, itemsep=5pt]
    \item \textbf{Turn 1 - Solve} - Given a question, provide a code solution to solve the provided question.
    \item  \textbf{Turn 2 - Simulate} - Given a code snippet and a test input, simulate the execution and predict the output that will be printed to the standard output. This step is performed independently for each public test.
    \item \textbf{Turn 3 - Submit or Fix} - Given a question, a candidate solution and feedback about each test (input, expected output, predicted output), decide whether the code is correct or not. If correct, submit the code solution, otherwise, fix the code to provide a new solution.
    \item \textbf{Optional} - Repeat turns 2 and 3 until a code solution is submitted or the maximum turns are reached.
\end{itemize}

Since the model’s ability to accurately predict execution outcomes may be weak at the start of RL training, relying solely on self-predicted feedback could lead the model to disregard this noisy signal. To mitigate this, we initially provide ground-truth execution feedback during training. As training progresses, one might switch from true execution signals to model-predicted execution outputs~\citep{bengio2015scheduled}. Alternatively, transition can also be deferred entirely to inference time. Our preliminary results showed no noticeable gap between the approaches, so we use the latter for simplicity. We denote the following approach \emph{Self-RLEF}. 
\section{Experimental Setup}

\subsection{Datasets}
We describe all datasets and configurations used to train and evaluate our models and baselines below. Note that each problem in competitive programming usually includes between one and four public test cases, typically provided in the problem description. These serve as basic checks for correctness and output formatting. In addition, a larger set of private tests, unavailable to the model, is used to better assess solution correctness, including coverage of edge cases and compliance with runtime constraints.

\subsubsection{Training datasets}
\label{sec:train_data}
The \nlex dataset, as presented in~\cref{sec:nlex}, was used for supervised fine-tuning, together with OpenMathReasoning~\cite{omr} and OpenCodeReasoning~\cite{ocr} datasets, to bootstrap reasoning abilities. 

During RL, models were optimised for both solving and predicting the output of competitive programming solutions. For that, we use the training split derived from CodeContests~\citep{dmc}, after heuristically filtering malformed instances, which results in ${\sim}12.2$k problems. For output prediction, we sample $10$ candidate solutions and retain up to four correct ones per model, since all correct submissions yield identical outputs. We then pair each retained solution with all of its public test cases, treating each as an output prediction instance, resulting in a total of ${\sim}143k$ code–input–output examples. All solutions were generated using CWM and Qwen2.5-7B \cite{qwen25}. 

\begin{figure}[t!]
  \begin{center}
    \centerline{\includegraphics[width=0.9\columnwidth]{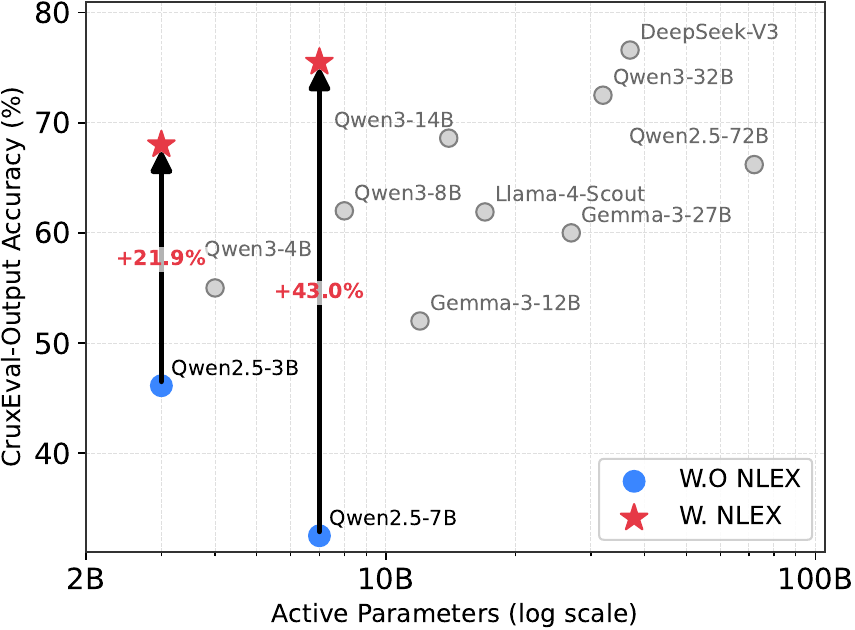}}
    \caption{
      \cruxo performance compared to model active parameters. Arrows demonstrate the benefit from training on \nlex data. We also compare to open models.
    }
    \label{fig:nl_tracing_sft}
  \end{center}  
\end{figure}

\subsubsection{Evaluation datasets}
We present evaluation datasets for competitive programming questions (first two), and output prediction (last two).

\newpara{\lcbio.} We curate a subset from LiveCodeBench-v6~\cite{lcb} containing only problems evaluated via \texttt{stdio} tests, which we refer to as \lcbio. This restriction simplifies output prediction, as the task reduces to determining the content written to \texttt{stdout} given a specific \texttt{stdin}. The resulting set includes $287$ problems. 

\newpara{\dmcvt.} We follow \citet{rlef} and use the validation and test splits of CodeContests~\citep{dmc}, yielding an additional evaluation set with a different distribution, denoted \dmcvt, and consisting of $282$ problems. 

\newpara{\cruxo}~\citep{cruxeval} is a widely adopted benchmark consisting of short Python functions paired with input–output examples. The task requires the model to infer the function’s return value given the code and its input. 

\newpara{Output prediction for competitive programming.} We generate $20$ solutions per-question from both \dmcvt and \lcbio using the same LLMs as mentioned above, without filtering or de-duplicating solutions, to perfectly match the real distribution of generated solutions. Such data is also used for best@k type metric calculations.

\subsection{Trained Models}
\label{sub:models}

We post-train Qwen2.5-Base models of sizes $3$B and $7$B, together with CWM-base using the datasets described in~\cref{sec:train_data}. For RL we use an asynchronous RL infrastructure, adopting the same RL algorithm as in CWM, with different hyperparameters. When performing multi-task training we employ sample-level weighting. Furthermore, we apply reward scaling, following \citet{critiquecoder}, assigning a weight of $0.8$ to the output prediction objective. For all multi-turn repair environments, including self-RLEF, we allow a maximum of one repair attempt during training (two solution turns in total, including the initial attempt), and $9$ at inference (overall $10$ turns). Full training configurations, including hyperparameters, are provided in Appendix~\ref{app:hps}.
\begin{figure*}[t!]
  \begin{center}
\centerline{\includegraphics[width=\textwidth]{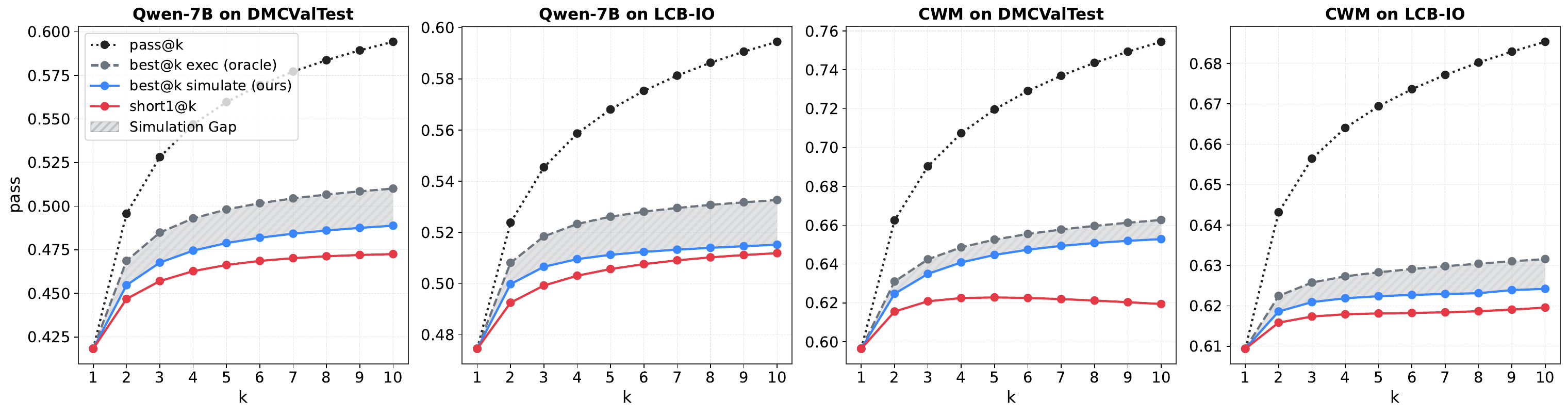}}
    \caption{Best@k performance of \emph{self-verification} with \emph{self-simulation}. Solutions and output predictions are produced by the same model - based on Qwen2.5-7B or CWM, trained for both solving and output prediction. Even though the tests used for filtering are in the solve prompt, there is still room for notable gains from simulating them.}
    \label{fig:self_verification}
  \end{center}
\end{figure*}

\begin{table}[t!]
  \caption{Output prediction performance of Qwen models trained with RLVR for output prediction, with and without \nlex data.}
  \label{tab:out_pred_results}
  \begin{center}
    \begin{small}
      \begin{sc}
      \resizebox{\columnwidth}{!}{
        \begin{tabular}{lcccccc}
          \toprule
          Base Model & \multicolumn{3}{c}{\lcbio-Out} & \multicolumn{3}{c}{DMC-Test-Out} \\
                & pass@1 & @5 & @10 & @1 & @5 & @10 \\
          \midrule
          Qwen-7B (ours) & \bf{79.7} & \bf{89.4} & \bf{91.1} & \bf{77.3} & \bf{89.3} & \bf{91.6} \\
          \quad w.o \nlex & 75.7 & 87.8 & 89.7 & 71.8 & 87.0 & 89.9 \\          
          \midrule
          Qwen-3B (ours) & \bf{66.4} & \bf{80.6} & \bf{83.9} & \bf{59.4} & \bf{78.2} & \bf{82.8} \\
          \quad w.o \nlex & 57.1 & 74.2 & 78.3 & 45.9 & 66.2 & 72.4 \\
          \bottomrule
        \end{tabular}}
      \end{sc}
    \end{small}
  \end{center}
  \vskip -0.1in
\end{table}

\section{Results}
\label{sec:res}

\subsection{Output Prediction}
\label{sec:res_output_pred}

\newpara{\cruxo.} We start by evaluating performance considering \cruxo. Results are presented in~\cref{fig:nl_tracing_sft}. We evaluate both Qwen2.5-3B and Qwen2.5-7B (after SFT only), trained with and without the \nlex data. For reference we provide pass@1 scores of common open-weights LLMs. Results show a clear superiority of the \nlex data as part of the training mix, achieving comparable performance to significantly larger models, with Qwen2.5-3B increasing from $37.5$ to $68.0$ and Qwen2.5-7B improving $48.5$ to $75.5$ pass@1 scores. We provide results for standard coding metrics in Appendix~\ref{app:results}, showing no regression in performance considering other benchmarks and tasks.

\newpara{Competitive programming.} Next, we evaluate output prediction performance on LLM solutions to competitive programming questions from \lcbio and \dmcvt (test split). Compared to \cruxo, these functions are often more complex and challenging. For that, we consider post-trained Qwen2.5 models (3B and 7B) on the task of output-prediction. Similarly to before, we consider models trained with and without \nlex data as part of the mix. Results presented in Table \ref{tab:out_pred_results} suggest that including \nlex data as part of the mix boosts output-prediction capabilities also after RL. While RL on output prediction with a standard reasoning SFT data (i.e., OMR and OCR) shows impressive performance, mixing them with \nlex provides superior results across both model sizes. To understand the effect of the RL phase, we additionally evaluate CWM on output prediction with and without RL. As expected, the RL phase significantly improves results on output prediction, see Appendix \ref{app:rl_abb} and Table \ref{tab:out_pred_rl_abb} for more details.

\newpara{Self-execution prediction.} So far we evaluated output-prediction capabilities on code generated by external models. We now turn to evaluate \emph{self-execution}, i.e. models perform output-prediction on their own generated solutions. For that we post-train CWM and Qwen2.5-7B on both output prediction and competitive problems solving. We report results for questions derived from both \lcbio and \dmcvt in Table \ref{tab:self_out_pred} and compare performance to models trained to perform output-prediction only (as a topline) and to the official CWM model (as a baseline). As expected, results suggest that jointly training both models for solving competitive programming questions and output prediction perform worse than output prediction only. For instance, CWM reaches $80.2$ and $86.5$ pass@1 scores in joint training compared to $85.0$ and $88.6$ scores when trained for output prediction only. However, both are significantly superior to the official CWM model, that reaches $57.7$ pass@1. Interestingly, these results suggest that unlike previous findings~\citep{gu2024counterfeit} models can reliably perform self-execution. 

\begin{table}[t!]
  \caption{Output prediction performance for models trained on standard code solving, jointly with output prediction (Joint), on their own solutions. We compare this to a model trained for output prediction  only, models from Tab. \ref{tab:out_pred_results}, (Out Pred), and official CWM. \label{tab:self_out_pred}}
  \begin{center}
    \begin{small}
      \begin{sc}
        \resizebox{\columnwidth}{!}{
        \begin{tabular}{llcccc}
          \toprule
          Model & RL obj. & \multicolumn{2}{c}{\dmcvt-Out}   & \multicolumn{2}{c}{\lcbio-Out}  \\
                  & & @1 & @5 & @1 & @5 \\
          \midrule
          CWM & official & 57.7 & 80.4 & 68.6 & 87.9  \\          
          & Joint & 80.2 & 87.2 & 86.5 & 91.0  \\
          & Out Pred & \textbf{85.0} & \textbf{89.8} & \textbf{88.6} & \textbf{92.7}  \\
          \midrule
          Qwen-7B       & Joint & 68.4 & 83.1 & 76.5 & 87.1  \\
          & Out Pred & \textbf{74.6} & \textbf{86.8} & \textbf{80.1} & \textbf{89.2}  \\
          \bottomrule
        \end{tabular}}
      \end{sc}
    \end{small}
  \end{center}  
  \vskip -0.1in
\end{table}

\subsection{Self-Execution for Competitive Programming}
\label{sec:res_self}

\begin{table*}[t!]
  \caption{Solve rates for training and evaluating with a standard reasoning approach vs using real or simulated execution feedback.}
  \label{tab:ef_solving}
  \begin{center}
    \begin{small}
      \begin{sc}
        \begin{tabular}{l|ccc|c|ccc|c}
          \toprule
           & \multicolumn{4}{c}{\dmcvt}   & \multicolumn{4}{c}{\lcbio}  \\
          & pass@1 & pass@5 & pass@10 & public & pass@1 & pass@5 & pass@10 & public \\
          \midrule
          Execution RLEF (Oracle) & 65.3 & 77.6 & 80.6 & 86.1 & 63.8 & 70.9 & 72.8 & 88.5 \\
          \midrule
          \midrule
          CWM (official)     & 49.0 & 63.7 & 67.9 & 60.8 & 57.4 & 67.3 & 70.1 & 71.4 \\
          CWM RL & 60.8 & 72.8 & 76.0 & 74.7 & 61.0 & 67.6 & 69.2 & 82.9 \\
          Self-RLEF (Ours) & \bf{63.2} & \bf{76.8} & \bf{80.2} & \bf{82.5} & \bf{62.3} & \bf{70.0} & \bf{71.9} & \bf{87.1} \\
          \bottomrule
        \end{tabular}
      \end{sc}
    \end{small}
  \end{center}  
\end{table*}

\paragraph{Self-verification.} Given a model's prediction of the execution output of its own code on public tests, one can use this to self-verify the solutions. Specifically, following the \textit{best@k simulate} approach described in \cref{sec:solving_with_sim}, we select and submit the solution predicted to pass most tests. To better estimate the effectiveness of the proposed method, we compare it to short1@k~\cite{michael_short}, which selects the shortest response among the $k$ solutions, and pass@k (for reference). To directly assess the quality of execution simulation, we also compare against an oracle that executes the public tests, following the same filtering procedure (denoted \textit{best@k exec}). This comparison will provide us with the \emph{simulation gap}, i.e., the performance gap between fully executing the code vs. simulating it with the model.
Our results provided in Fig.~\ref{fig:self_verification} show that self-verification provides a large boost in performance under the best@k setup, ($2-8$ points compared to standard solving), \emph{despite the tests used for filtering being provided when generating the solution}. This also outperforms short1@k. Notably, for Qwen2.5-7B the simulation gap is larger than CWM perhaps implying the need for larger or stronger models to learn to both solve and simulate execution effectively. Further results in Fig.~\ref{fig:qwen_solve_verification_qwen_out} show that using Qwen2.5-7B trained for output prediction only to filter the same solutions leads to a smaller gap.

\paragraph{Self-RLEF.} We train our model using the procedure described in \cref{sec:multi-turn}. We report pass@k scores for $k \in \{1, 5, 10\}$ on both \lcbio and \dmcvt, along with public-test pass rates. We evaluate three variants: the official CWM model, CWM post-trained specifically for competitive programming (CWM-RL), and CWM jointly optimised for output prediction and competitive programming with execution feedback. The latter is evaluated under the proposed self-RLEF framework, using either simulated execution or real execution as an oracle. In both settings, the model is allowed up to $10$ coding turns (initial solution + $9$ fix), although in practice, it uses $3.33$ turns on average (Appendix \ref{app:srlef} provides additional results with less turns). Results in \cref{tab:ef_solving} show that self-RLEF consistently outperforms both the official CWM and CWM-RL across all settings, improving pass rates on both public and private tests. Compared to the oracle, a performance gap remains, particularly for pass@1, indicating room for improvement in execution prediction. Interestingly, pass@1 scores (with 10 turns) are lower than the corresponding best@10 results shown in \cref{fig:self_verification}. We hypothesise that this gap arises from limited exploration, as the model tends to iteratively fix a solution rather than explore alternative ones. In addition, the model frequently does not use all turns as seen by the average number of turns. For certain settings where an "early exit" is preferred this approach can provide a better tradeoff considering compute.

\subsection{Ablations}\label{sec:res_ablations}

\begin{table}[t!]
  \caption{Comparing performance of using the self-RLEF scaffold at inference only with open source reasoning models. }
  \label{tab:rlef_scaf}
  \begin{center}
    \begin{small}
    \begin{sc}
    \resizebox{\columnwidth}{!}{
    \begin{tabular}{lcccccc}
        \toprule
        & \multicolumn{3}{c}{\dmcvt}   & \multicolumn{3}{c}{\lcbio}  \\
        Model \textbackslash~Pass      & @1 & @5 & @10 & @1 & @5 & @10 \\
        \midrule
        Qwen3-32B & 44.7 & 61.4 & 66.0 & 58.6 & 68.9 & 72.2 \\
        \quad + SRLEF ($\Delta$) & -10.6 & -2.2 & -1.4 & -20.1 & -1.0 & -0.1 \\
        \midrule
        CWM             & 49.0 & 63.7 & 67.9 & 57.4 & 67.3 & 70.1 \\
        \quad + SRLEF ($\Delta$) & -4.8 & +0.5 & +0.9 & -7.4 & +0.1 & +0.2 \\
        \bottomrule
    \end{tabular}}
    \end{sc}
    \end{small}
    \end{center}
  \vskip -0.1in
\end{table}

\newpara{Self-RLEF scaffold.} One may wonder to what extent the self-refinement pipeline itself leads to the inference performance gain irrespective of model training. Hence, we investigate inference using the Self-RLEF approach with public open-weights models, specifically Qwen3-32B and CWM. We compare these results to using these models in a standard single turn inference procedure.
Results provided in \cref{tab:rlef_scaf} show no noticeable improvement from using the proposed self-RLEF approach, and even a decrease in performance across both models over all metrics and datasets. By manual analysis, we observe the model struggles to correctly predict the output, and ignores the feedback. 

\begin{figure*}[t!]
  \begin{center}    
  \centerline{\includegraphics[width=0.88\textwidth]{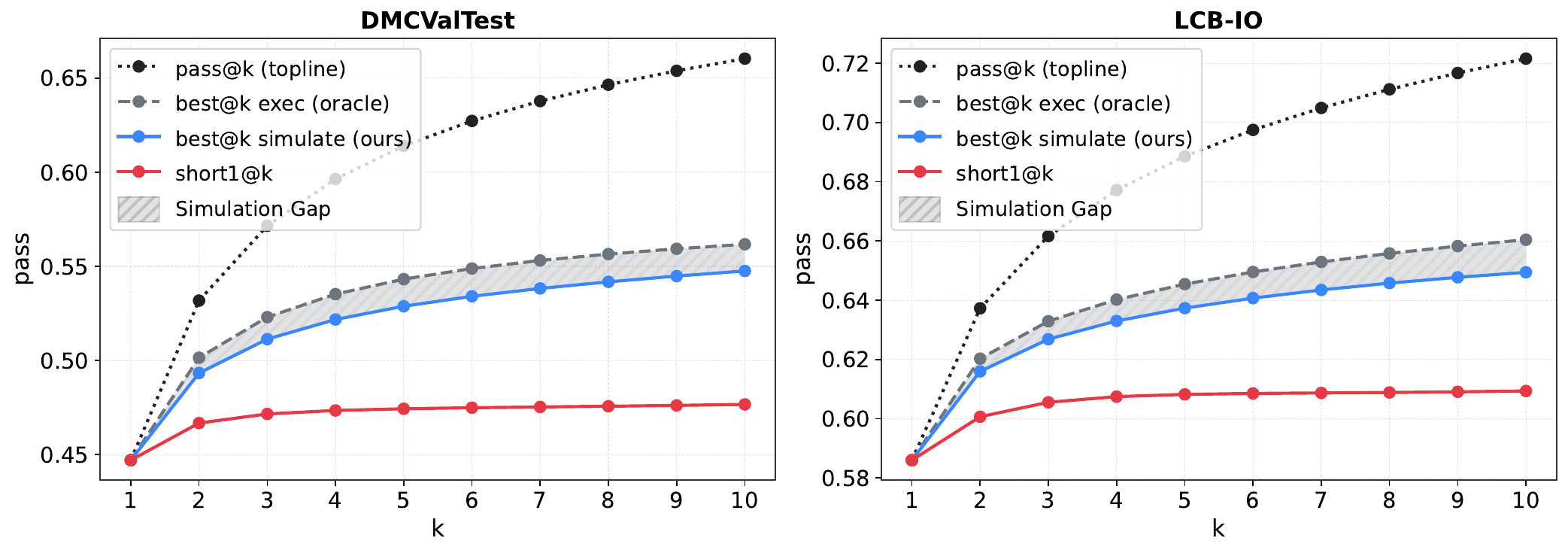}}
    \caption{Comparing best@k when ranking Qwen3-32B solutions, using CWM post-trained only for output prediction as a verifier. }
    \label{fig:qwen32_verification_cwm}
  \end{center}
\end{figure*}

\newpara{Knowing when to submit or fix.} To better identify the source of performance gains from self-RLEF, we analyse model behaviour during inference time. We measure how often the model successfully fixes a solution, i.e., cases where the initial solution fails the tests but the final solution passes, as well as how often it breaks a solution, i.e., where an initially passing solution fails after revision. For completeness, we also report cases where both the initial and final solutions either pass or fail the tests. \cref{tab:fixing_confusion} reports results for CWM on \dmcvt using both public and private tests. Results suggest the model rarely degrades correct solutions, breaking only $1.2\%$ of cases on public tests and $5.0\%$ on private tests. In contrast, when the initial solution fails, the model frequently produces effective fixes, succeeding in $17.0\%$ of public-tests and $10.4\%$ of private-tests. These improvements increase the pass rate from $57.8\%$ for the initial solutions to $63.2\%$ for the final submissions.

\begin{table}[t!]
    \centering
    \caption{Pass rates of the initial generated solution (Init), compared to the final submitted solution  (Sub) in \emph{Self-RLEF} inference on \dmcvt considering both public and private tests. }
    \label{tab:fixing_confusion}

    \begin{subtable}{0.45\columnwidth}
    \small
        \centering
        \caption{Public Tests}
        \begin{tabular}{lcc}
            \toprule
            Init\textbackslash Sub & Fail & Pass \\
            \midrule
            Fail  & 16.3\%  & 17.0\%  \\
            Pass  & 1.2\%   & 65.5\%  \\
            \bottomrule
        \end{tabular}
    \end{subtable}
    \hspace{1em}
    \begin{subtable}{0.45\columnwidth}
    \small
        \centering
        \caption{Private Tests}
        \begin{tabular}{lcc}
            \toprule
            Init\textbackslash Sub & Fail & Pass \\
            \midrule
            Fail  & 31.8\%  & 10.4\%  \\
            Pass  & 5.0\%   & 52.8\%  \\
            \bottomrule
        \end{tabular}
    \end{subtable}
\end{table}

\subsection{Beyond \emph{Self}-Verification}
\label{sec:res_beyond}
While the focus of this work is \emph{self}-execution, one could imagine use cases for building a model as a code simulator only. As an initial analysis, we measure public test pass rates for multiple models on \lcbio and \dmcvt to assess the potential of the best@k approach. \cref{tab:pub_pass} reports public pass@1 and pass@10 results for Qwen and CWM. While models can generate solutions which pass the public tests (as evidenced in pass@10), they often submit solutions which do not pass them \emph{even though they are provided in the question}. This suggests that standard reasoning approaches under-utilise such test information. This observation motivates the use of solution verification.

We next assess the ability of our trained CWM model to predict execution outputs for solutions generated by Qwen3-32B. CWM achieves pass@1 and pass@5 scores of $86.1$ and $91.4$, respectively, on public test output prediction. Based on this, we apply the best@k evaluation on both \lcbio and \dmcvt, as shown in \cref{fig:qwen32_verification_cwm}. The results indicate that using CWM with this filtering strategy is very effective and can correctly filter solutions generated by external models. Furthermore, compared to real execution, we observe only a minor \emph{simulation gap}. This again highlights the efficacy of the output prediction method to alleviate the need for execution, and further shows generalisation to other models' solutions. Results for this setup using Qwen3-4B and CWM-RL are provided in Appendix \ref{app:results}, showing similar trends. 

\begin{table}[t!]
  \caption{Public pass@1 and 10 of different models. The large gap of standard reasoning models can suggest that they under-utilise provided test information.}
  \label{tab:pub_pass}
  \begin{center}
    \begin{small}
      \begin{sc}
        \resizebox{0.85\columnwidth}{!}{
        \begin{tabular}{lcccc}
          \toprule
          & \multicolumn{2}{c}{\dmcvt}   & \multicolumn{2}{c}{\lcbio}  \\
          Model \textbackslash~Public Pass      & @1 & @10 & @1 & @10 \\
          \midrule
          Qwen3-4B  & 42.5 & 65.4 & 64.4 & 80.9  \\
          Qwen3-32B & 56.3 & 79.0 & 72.3 & 88.8  \\
          Qwen2.5-7B RL & 55.1 & 76.5 & 68.0 & 84.7  \\
          CWM RL     & 73.4 & 87.6 & 81.7 & 90.6  \\
          \bottomrule
        \end{tabular}}
      \end{sc}
    \end{small}
  \end{center}
\end{table}
\section{Related Work}
\paragraph{Code Simulation \& Verification.}
Several works ask how well LLMs can simulate or predict the output of a given code snippet \cite{hora2024predicting, licodeio, cruxeval, cruxevalx, cwm, armengol2025cannot}. Others suggest that models struggle to simulate their own code, as they are blind to its flaws \cite{gu2024counterfeit}. Some works use models to simulate tool execution as part of a synthetic data generation \cite{kimi}. Furthermore, some studies explicitly train a model as a verifier of solutions \cite{le2022coderl}. Many report challenges in verification performance \cite{critiquecoder, critique_ft}.

\paragraph{Learning from Feedback.} \citet{rlef} showed that models can learn to utilise feedback about the execution of their generated code. 
Providing models with access to interpreters is a popular approach that has been used to improve performance in maths \cite{chen2023program, pmlr-v202-gao23f}, code generation \cite{liu2023dynamic, shinn2023reflexion}, competitive programming \cite{zheng2025what}, and agentic coding \cite{swe_agent, xia2025live}. Several prompting approaches were suggested for non-reasoning models to elicit self-improvement, often joint with external verification signals \citep{chen2023teaching, renze2024self, madaan2023self, kumar2024training}. \citet{chen2023improving} further showed that training with human written feedback on code can improve performance.
\section{Discussion}
\newpara{Limitations.}
The main limitation of simulating program execution is estimating complex computational operations (e.g., multiplying large numbers, compute logarithms, etc.). Yet, while execution simulation is imperfect and can introduce noise, our findings suggest that it provides a useful inductive bias for reasoning about program behaviour, particularly in settings where direct execution is expensive or infeasible. Furthermore, while our approach showed promising results, it is currently limited to single file competitive programming questions. Generalising this to full repository SWE tasks poses an interesting future research direction.

\newpara{Future Work.} We believe our work opens several directions for future research. The most interesting direction in our opinion is using the full rich execution simulation, and not only the final output as feedback for iterative code fixing. Such feedback can convey richer information than the output alone (\emph{even beyond real execution}), capturing not just what output is produced, but why it arises. Such explanations can reveal cases where a test appears to pass for incidental reasons, as well as provide insight into the underlying causes of failures. In preliminary results we observe that training with rich textual feedback presents challenges to training stability. We hypothesise this is due to both inability to train with teacher forcing and unclear definition of the verifiable reward of the simulation. We leave such exploration for future work. 

\section{Conclusion}
In this work we investigated if LLMs can be trained to simulate code execution and whether this capability can be used to improve code generation. By combining SFT on \nlex with RLVR, we showed that models can acquire the ability to predict execution outcomes for general programs as well as code they generate. Leveraging this ability, we introduced \emph{self-verification} and iterative \emph{self-fix} strategies using predicted execution signals to select or refine candidate solutions without relying on external execution. Our empirical results on competitive programming tasks demonstrate consistent improvements over standard baselines considering both output prediction and question solving. Compared with real execution we  show a relatively small \emph{simulation gap}, demonstrating the usability of the proposed approach compared to the top-line of code execution. More broadly, our results suggest that enabling models to reason about the outcomes of the code they generate may be a key for building more reliable programming agents.

% \section*{Acknowledgements}

% \textbf{Do not} include acknowledgements in the initial version of the paper

% If a paper is accepted, the final camera-ready version can (and usually should)
% include acknowledgements.  Such acknowledgements should be placed at the end of
% the section, in an unnumbered section that does not count towards the paper
% page limit. Typically, this will include thanks to reviewers who gave useful
% comments, to colleagues who contributed to the ideas, and to funding agencies
% and corporate sponsors that provided financial support.

\section*{Impact Statement}
This paper presents work whose goal is to advance the field of Machine Learning. There are many potential societal consequences of our work, none which we feel must be specifically highlighted here.

\bibliography{icml}

@inproceedings{
shinn2023reflexion,
title={Reflexion: language agents with verbal reinforcement learning},
author={Noah Shinn and Federico Cassano and Ashwin Gopinath and Karthik R Narasimhan and Shunyu Yao},
booktitle={Thirty-seventh Conference on Neural Information Processing Systems},
year={2023},
url={https://openreview.net/forum?id=vAElhFcKW6}
}

@article{
chen2023program,
title={Program of Thoughts Prompting: Disentangling Computation from Reasoning for Numerical Reasoning Tasks},
author={Wenhu Chen and Xueguang Ma and Xinyi Wang and William W. Cohen},
journal={Transactions on Machine Learning Research},
issn={2835-8856},
year={2023},
url={https://openreview.net/forum?id=YfZ4ZPt8zd},
note={}
}

@inproceedings{
zheng2025what,
title={What Makes Large Language Models Reason in (Multi-Turn) Code Generation?},
author={Kunhao Zheng and Juliette Decugis and Jonas Gehring and Taco Cohen and benjamin negrevergne and Gabriel Synnaeve},
booktitle={The Thirteenth International Conference on Learning Representations},
year={2025},
url={https://openreview.net/forum?id=Zk9guOl9NS}
}

@InProceedings{pmlr-v202-gao23f,
  title = 	 {{PAL}: Program-aided Language Models},
  author =       {Gao, Luyu and Madaan, Aman and Zhou, Shuyan and Alon, Uri and Liu, Pengfei and Yang, Yiming and Callan, Jamie and Neubig, Graham},
  booktitle = 	 {Proceedings of the 40th International Conference on Machine Learning},
  pages = 	 {10764--10799},
  year = 	 {2023},
  editor = 	 {Krause, Andreas and Brunskill, Emma and Cho, Kyunghyun and Engelhardt, Barbara and Sabato, Sivan and Scarlett, Jonathan},
  volume = 	 {202},
  series = 	 {Proceedings of Machine Learning Research},
  month = 	 {23--29 Jul},
  publisher =    {PMLR},
  url = 	 {https://proceedings.mlr.press/v202/gao23f.html}
}

@misc{liu2023dynamic,
    title={Dynamic LLM-Agent Network: An LLM-agent Collaboration Framework with Agent Team Optimization},
    author={Zijun Liu and Yanzhe Zhang and Peng Li and Yang Liu and Diyi Yang},
    year={2023},
    eprint={2310.02170},
    archivePrefix={arXiv},
    primaryClass={cs.CL}
}

@InProceedings{cruxeval,
  title = 	 {{CRUXE}val: A Benchmark for Code Reasoning, Understanding and Execution},
  author =       {Gu, Alex and Roziere, Baptiste and Leather, Hugh James and Solar-Lezama, Armando and Synnaeve, Gabriel and Wang, Sida},
  booktitle = 	 {Proceedings of the 41st International Conference on Machine Learning},
  pages = 	 {16568--16621},
  year = 	 {2024},
  editor = 	 {Salakhutdinov, Ruslan and Kolter, Zico and Heller, Katherine and Weller, Adrian and Oliver, Nuria and Scarlett, Jonathan and Berkenkamp, Felix},
  volume = 	 {235},
  series = 	 {Proceedings of Machine Learning Research},
  month = 	 {21--27 Jul},
  publisher =    {PMLR},
  url = 	 {https://proceedings.mlr.press/v235/gu24c.html},
}

@article{armengol2025cannot,
  title={What I cannot execute, I do not understand: Training and Evaluating LLMs on Program Execution Traces},
  author={Armengol-Estap{\'e}, Jordi and Carbonneaux, Quentin and Zhang, Tianjun and Markosyan, Aram H and Seeker, Volker and Cummins, Chris and Kambadur, Melanie and O'Boyle, Michael FP and Wang, Sida and Synnaeve, Gabriel and others},
  journal={arXiv preprint arXiv:2503.05703},
  year={2025}
}

@inproceedings{rlef,
title={{RLEF}: Grounding Code {LLM}s in Execution Feedback with Reinforcement Learning},
author={Jonas Gehring and Kunhao Zheng and Jade Copet and Vegard Mella and Taco Cohen and Gabriel Synnaeve},
booktitle={Forty-second International Conference on Machine Learning},
year={2025},
url={https://openreview.net/forum?id=PzSG5nKe1q}
}

@INPROCEEDINGS{11052794,
  author={Peng, Yun and Gotmare, Akhilesh Deepak and Lyu, Michael R. and Xiong, Caiming and Savarese, Silvio and Sahoo, Doyen},
  booktitle={2025 IEEE/ACM Second International Conference on AI Foundation Models and Software Engineering (Forge)}, 
  title={PerfCodeGen: Improving Performance of LLM Generated Code with Execution Feedback}, 
  year={2025},
  volume={},
  number={},
  pages={1-13},
  doi={10.1109/Forge66646.2025.00008}}

@article{swe_agent,
  title={Swe-agent: Agent-computer interfaces enable automated software engineering},
  author={Yang, John and Jimenez, Carlos E and Wettig, Alexander and Lieret, Kilian and Yao, Shunyu and Narasimhan, Karthik and Press, Ofir},
  journal={Advances in Neural Information Processing Systems},
  volume={37},
  pages={50528--50652},
  year={2024}
}

@inproceedings{silver_bullet,
title={Is Self-Repair a Silver Bullet for Code Generation?},
author={Theo X. Olausson and Jeevana Priya Inala and Chenglong Wang and Jianfeng Gao and Armando Solar-Lezama},
booktitle={The Twelfth International Conference on Learning Representations},
year={2024},
url={https://openreview.net/forum?id=y0GJXRungR}
}

@inproceedings{cruxevalx,
    title = "{CRUXEVAL}-{X}: A Benchmark for Multilingual Code Reasoning, Understanding and Execution",
    author = "Xu, Ruiyang  and
      Cao, Jialun  and
      Lu, Yaojie  and
      Wen, Ming  and
      Lin, Hongyu  and
      Han, Xianpei  and
      He, Ben  and
      Cheung, Shing-Chi  and
      Sun, Le",
    editor = "Che, Wanxiang  and
      Nabende, Joyce  and
      Shutova, Ekaterina  and
      Pilehvar, Mohammad Taher",
    booktitle = "Proceedings of the 63rd Annual Meeting of the Association for Computational Linguistics (Volume 1: Long Papers)",
    month = jul,
    year = "2025",
    address = "Vienna, Austria",
    publisher = "Association for Computational Linguistics",
    url = "https://aclanthology.org/2025.acl-long.1158/",
    doi = "10.18653/v1/2025.acl-long.1158",
    pages = "23762--23779",
    ISBN = "979-8-89176-251-0",
}

@inproceedings{gu2024counterfeit,
  title={The Counterfeit Conundrum: Can Code Language Models Grasp the Nuances of Their Incorrect Generations?},
  author={Gu, Alex and Li, Wen-Ding and Jain, Naman and Olausson, Theo and Lee, Celine and Sen, Koushik and Solar-Lezama, Armando},
  booktitle={Findings of the Association for Computational Linguistics ACL 2024},
  pages={74--117},
  year={2024}
}

@article{kamoi2024can,
  title={When Can LLMs Actually Correct Their Own Mistakes? A Critical Survey of Self-Correction of LLMs},
  author={Kamoi, Ryo and Zhang, Yusen and Zhang, Nan and Han, Jiawei and Zhang, Rui},
  journal={Transactions of the Association for Computational Linguistics},
  volume={12},
  pages={1417--1440},
  year={2024}
}

@article{dmc,
  title={Competition-level code generation with alphacode},
  author={Li, Yujia and Choi, David and Chung, Junyoung and Kushman, Nate and Schrittwieser, Julian and Leblond, R{\'e}mi and Eccles, Tom and Keeling, James and Gimeno, Felix and Dal Lago, Agustin and others},
  journal={Science},
  volume={378},
  number={6624},
  pages={1092--1097},
  year={2022},
  publisher={American Association for the Advancement of Science}
}

@article{yang2025qwen3,
  title={Qwen3 technical report},
  author={Yang, An and Li, Anfeng and Yang, Baosong and Zhang, Beichen and Hui, Binyuan and Zheng, Bo and Yu, Bowen and Gao, Chang and Huang, Chengen and Lv, Chenxu and others},
  journal={arXiv preprint arXiv:2505.09388},
  year={2025}
}

@misc{qwen25,
      title={Qwen2.5 Technical Report}, 
      author={Qwen and : and An Yang and Baosong Yang and Beichen Zhang and Binyuan Hui and Bo Zheng and Bowen Yu and Chengyuan Li and Dayiheng Liu and Fei Huang and Haoran Wei and Huan Lin and Jian Yang and Jianhong Tu and Jianwei Zhang and Jianxin Yang and Jiaxi Yang and Jingren Zhou and Junyang Lin and Kai Dang and Keming Lu and Keqin Bao and Kexin Yang and Le Yu and Mei Li and Mingfeng Xue and Pei Zhang and Qin Zhu and Rui Men and Runji Lin and Tianhao Li and Tianyi Tang and Tingyu Xia and Xingzhang Ren and Xuancheng Ren and Yang Fan and Yang Su and Yichang Zhang and Yu Wan and Yuqiong Liu and Zeyu Cui and Zhenru Zhang and Zihan Qiu},
      year={2025},
      eprint={2412.15115},
      archivePrefix={arXiv},
      primaryClass={cs.CL},
      url={https://arxiv.org/abs/2412.15115}, 
}

@article{llama3,
  title={The llama 3 herd of models},
  author={Grattafiori, Aaron and Dubey, Abhimanyu and Jauhri, Abhinav and Pandey, Abhinav and Kadian, Abhishek and Al-Dahle, Ahmad and Letman, Aiesha and Mathur, Akhil and Schelten, Alan and Vaughan, Alex and others},
  journal={arXiv preprint arXiv:2407.21783},
  year={2024}
}

@article{cwm,
  title={Cwm: An open-weights llm for research on code generation with world models},
  author={Copet, Jade and Carbonneaux, Quentin and Cohen, Gal and Gehring, Jonas and Kahn, Jacob and Kossen, Jannik and Kreuk, Felix and McMilin, Emily and Meyer, Michel and Wei, Yuxiang and others},
  journal={arXiv preprint arXiv:2510.02387},
  year={2025}
}

@article{critique_ft,
  title={Critique fine-tuning: Learning to critique is more effective than learning to imitate},
  author={Wang, Yubo and Yue, Xiang and Chen, Wenhu},
  journal={arXiv preprint arXiv:2501.17703},
  year={2025}
}

@article{critiquecoder,
  title={Critique-Coder: Enhancing Coder Models by Critique Reinforcement Learning},
  author={Ruan, Chi and Jiang, Dongfu and Wang, Yubo and Chen, Wenhu},
  journal={arXiv preprint arXiv:2509.22824},
  year={2025}
}

@article{larger,
  title={The larger the better? improved llm code-generation via budget reallocation},
  author={Hassid, Michael and Remez, Tal and Gehring, Jonas and Schwartz, Roy and Adi, Yossi},
  journal={arXiv preprint arXiv:2404.00725},
  year={2024}
}

@article{jimenez2023swe,
  title={Swe-bench: Can language models resolve real-world github issues?},
  author={Jimenez, Carlos E and Yang, John and Wettig, Alexander and Yao, Shunyu and Pei, Kexin and Press, Ofir and Narasimhan, Karthik},
  journal={arXiv preprint arXiv:2310.06770},
  year={2023}
}

@article{lcb,
  title={Livecodebench: Holistic and contamination free evaluation of large language models for code},
  author={Jain, Naman and Han, King and Gu, Alex and Li, Wen-Ding and Yan, Fanjia and Zhang, Tianjun and Wang, Sida and Solar-Lezama, Armando and Sen, Koushik and Stoica, Ion},
  journal={arXiv preprint arXiv:2403.07974},
  year={2024}
}

@article{omr,
  title={Aimo-2 winning solution: Building state-of-the-art mathematical reasoning models with openmathreasoning dataset},
  author={Moshkov, Ivan and Hanley, Darragh and Sorokin, Ivan and Toshniwal, Shubham and Henkel, Christof and Schifferer, Benedikt and Du, Wei and Gitman, Igor},
  journal={arXiv preprint arXiv:2504.16891},
  year={2025}
}

@article{ocr,
  title={Opencodereasoning: Advancing data distillation for competitive coding},
  author={Ahmad, Wasi Uddin and Narenthiran, Sean and Majumdar, Somshubra and Ficek, Aleksander and Jain, Siddhartha and Huang, Jocelyn and Noroozi, Vahid and Ginsburg, Boris},
  journal={arXiv preprint arXiv:2504.01943},
  year={2025}
}

@article{michael_short,
  title={Don't Overthink it. Preferring Shorter Thinking Chains for Improved LLM Reasoning},
  author={Hassid, Michael and Synnaeve, Gabriel and Adi, Yossi and Schwartz, Roy},
  journal={arXiv preprint arXiv:2505.17813},
  year={2025}
}

@inproceedings{yao2022react,
  title={React: Synergizing reasoning and acting in language models},
  author={Yao, Shunyu and Zhao, Jeffrey and Yu, Dian and Du, Nan and Shafran, Izhak and Narasimhan, Karthik R and Cao, Yuan},
  booktitle={The eleventh international conference on learning representations},
  year={2022}
}

@article{bengio2015scheduled,
  title={Scheduled sampling for sequence prediction with recurrent neural networks},
  author={Bengio, Samy and Vinyals, Oriol and Jaitly, Navdeep and Shazeer, Noam},
  journal={Advances in neural information processing systems},
  volume={28},
  year={2015}
}

@article{thimmaiah2025plsemanticsbench,
  title={PLSemanticsBench: Large Language Models As Programming Language Interpreters},
  author={Thimmaiah, Aditya and Zhang, Jiyang and Srinivasa, Jayanth and Li, Junyi Jessy and Gligoric, Milos},
  journal={arXiv preprint arXiv:2510.03415},
  year={2025}
}

@article{chen2023improving,
  title={Improving code generation by training with natural language feedback},
  author={Chen, Angelica and Scheurer, J{\'e}r{\'e}my and Korbak, Tomasz and Campos, Jon Ander and Chan, Jun Shern and Bowman, Samuel R and Cho, Kyunghyun and Perez, Ethan},
  journal={arXiv preprint arXiv:2303.16749},
  year={2023}
}

@article{chen2023teaching,
  title={Teaching large language models to self-debug},
  author={Chen, Xinyun and Lin, Maxwell and Sch{\"a}rli, Nathanael and Zhou, Denny},
  journal={arXiv preprint arXiv:2304.05128},
  year={2023}
}

@article{renze2024self,
  title={Self-reflection in llm agents: Effects on problem-solving performance},
  author={Renze, Matthew and Guven, Erhan},
  journal={arXiv preprint arXiv:2405.06682},
  year={2024}
}

@article{madaan2023self,
  title={Self-refine: Iterative refinement with self-feedback},
  author={Madaan, Aman and Tandon, Niket and Gupta, Prakhar and Hallinan, Skyler and Gao, Luyu and Wiegreffe, Sarah and Alon, Uri and Dziri, Nouha and Prabhumoye, Shrimai and Yang, Yiming and others},
  journal={Advances in Neural Information Processing Systems},
  volume={36},
  pages={46534--46594},
  year={2023}
}

@misc{minimax2026m21,
  author = {{MiniMax}},
  title = {M2.1: Multilingual and Multi-Task Coding with Strong Generalization},
  howpublished = {\url{https://x.com/MiniMax__AI/status/2007843119832695114}},
  year = {2026},
  month = {January 4}
}

@article{bogin2024super,
  title={Super: Evaluating agents on setting up and executing tasks from research repositories},
  author={Bogin, Ben and Yang, Kejuan and Gupta, Shashank and Richardson, Kyle and Bransom, Erin and Clark, Peter and Sabharwal, Ashish and Khot, Tushar},
  journal={arXiv preprint arXiv:2409.07440},
  year={2024}
}

@misc{mbpp,
  title={Program Synthesis with Large Language Models},
  author={Austin, Jacob and Odena, Augustus and Nye, Maxwell and Bosma, Maarten and Michalewski, Henryk and Dohan, David and Jiang, Ellen and Cai, Carrie and Terry, Michael and Le, Quoc and others},
  note={{arXiv}:2108.07732},
  url={https://arxiv.org/abs/2108.07732},
  year={2021}
}

@inproceedings{evalplus,
  title = {Is Your Code Generated by Chat{GPT} Really Correct? Rigorous Evaluation of Large Language Models for Code Generation},
  author = {Liu, Jiawei and Xia, Chunqiu Steven and Wang, Yuyao and Zhang, Lingming},
  booktitle = {Thirty-seventh Conference on Neural Information Processing Systems},
  year = {2023},
  url = {https://openreview.net/forum?id=1qvx610Cu7},
}

@article{math500,
      title={Let's Verify Step by Step}, 
      author={Lightman, Hunter and Kosaraju, Vineet and Burda, Yura and Edwards, Harri and Baker, Bowen and Lee, Teddy and Leike, Jan and Schulman, John and Sutskever, Ilya and Cobbe, Karl},
      journal={arXiv preprint arXiv:2305.20050},
      year={2023}
}

@article{gsm8k,
  title={Training Verifiers to Solve Math Word Problems},
  author={Cobbe, Karl and Kosaraju, Vineet and Bavarian, Mohammad and Chen, Mark and Jun, Heewoo and Kaiser, Lukasz and Plappert, Matthias and Tworek, Jerry and Hilton, Jacob and Nakano, Reiichiro and Hesse, Christopher and Schulman, John},
  journal={arXiv preprint arXiv:2110.14168},
  year={2021}
}

@article{xia2025live,
  title={Live-SWE-agent: Can Software Engineering Agents Self-Evolve on the Fly?},
  author={Xia, Chunqiu Steven and Wang, Zhe and Yang, Yan and Wei, Yuxiang and Zhang, Lingming},
  journal={arXiv preprint arXiv:2511.13646},
  year={2025}
}

@article{le2022coderl,
  title={Coderl: Mastering code generation through pretrained models and deep reinforcement learning},
  author={Le, Hung and Wang, Yue and Gotmare, Akhilesh Deepak and Savarese, Silvio and Hoi, Steven Chu Hong},
  journal={Advances in Neural Information Processing Systems},
  volume={35},
  pages={21314--21328},
  year={2022}
}

@article{kumar2024training,
  title={Training language models to self-correct via reinforcement learning},
  author={Kumar, Aviral and Zhuang, Vincent and Agarwal, Rishabh and Su, Yi and Co-Reyes, John D and Singh, Avi and Baumli, Kate and Iqbal, Shariq and Bishop, Colton and Roelofs, Rebecca and others},
  journal={arXiv preprint arXiv:2409.12917},
  year={2024}
}

@article{chan2024mle,
  title={Mle-bench: Evaluating machine learning agents on machine learning engineering},
  author={Chan, Jun Shern and Chowdhury, Neil and Jaffe, Oliver and Aung, James and Sherburn, Dane and Mays, Evan and Starace, Giulio and Liu, Kevin and Maksin, Leon and Patwardhan, Tejal and others},
  journal={arXiv preprint arXiv:2410.07095},
  year={2024}
}

@article{zheng2026can,
  title={Can We Predict Before Executing Machine Learning Agents?},
  author={Zheng, Jingsheng and Zhang, Jintian and Luo, Yujie and Mao, Yuren and Gao, Yunjun and Du, Lun and Chen, Huajun and Zhang, Ningyu},
  journal={arXiv preprint arXiv:2601.05930},
  year={2026}
}

@inproceedings{
licodeio,
title={Code{IO}: Condensing Reasoning Patterns via Code Input-Output Prediction},
author={Junlong Li and Daya Guo and Dejian Yang and Runxin Xu and Yu Wu and Junxian He},
booktitle={Forty-second International Conference on Machine Learning},
year={2025},
url={https://openreview.net/forum?id=feIaF6vYFl}
}

@inproceedings{hora2024predicting,
  title={Predicting test results without execution},
  author={Hora, Andre},
  booktitle={Companion Proceedings of the 32nd ACM International Conference on the Foundations of Software Engineering},
  pages={542--546},
  year={2024}
}

@article{ding2025understanding,
  title={Understanding world or predicting future? a comprehensive survey of world models},
  author={Ding, Jingtao and Zhang, Yunke and Shang, Yu and Zhang, Yuheng and Zong, Zefang and Feng, Jie and Yuan, Yuan and Su, Hongyuan and Li, Nian and Sukiennik, Nicholas and others},
  journal={ACM Computing Surveys},
  volume={58},
  number={3},
  pages={1--38},
  year={2025},
  publisher={ACM New York, NY}
}

@article{ha2018world,
  title={World models},
  author={Ha, David and Schmidhuber, J{\"u}rgen},
  journal={arXiv preprint arXiv:1803.10122},
  volume={2},
  number={3},
  year={2018}
}

@article{rope,
  title={Roformer: Enhanced transformer with rotary position embedding},
  author={Su, Jianlin and Ahmed, Murtadha and Lu, Yu and Pan, Shengfeng and Bo, Wen and Liu, Yunfeng},
  journal={Neurocomputing},
  volume={568},
  pages={127063},
  year={2024},
  publisher={Elsevier}
}

@article{qian2026current,
  title={Current Agents Fail to Leverage World Model as Tool for Foresight},
  author={Qian, Cheng and Acikgoz, Emre Can and Li, Bingxuan and Chen, Xiusi and Zhang, Yuji and He, Bingxiang and Luo, Qinyu and Hakkani-T{\"u}r, Dilek and Tur, Gokhan and Li, Yunzhu and others},
  journal={arXiv preprint arXiv:2601.03905},
  year={2026}
}

@article{kimi,
  title={Kimi k2: Open agentic intelligence},
  author={Kimi, Team and Bai, Yifan and Bao, Yiping and Chen, Guanduo and Chen, Jiahao and Chen, Ningxin and Chen, Ruijue and Chen, Yanru and Chen, Yuankun and Chen, Yutian and others},
  journal={arXiv preprint arXiv:2507.20534},
  year={2025}
}

@misc{beck2026neuraldebugger,
      title={Towards a Neural Debugger for Python}, 
      author={Maximilian Beck and Jonas Gehring and Jannik Kossen and Gabriel Synnaeve},
      year={2026},
      eprint={2603.09951},
      archivePrefix={arXiv},
      primaryClass={cs.LG},
      url={https://arxiv.org/abs/2603.09951}, 
}
\bibliographystyle{icml2026}

%%%%%%%%%%%%%%%%%%%%%%%%%%%%%%%%%%%%%%%%%%%%%%%%%%%%%%%%%%%%%%%%%%%%%%%%%%%%%%%
%%%%%%%%%%%%%%%%%%%%%%%%%%%%%%%%%%%%%%%%%%%%%%%%%%%%%%%%%%%%%%%%%%%%%%%%%%%%%%%
% APPENDIX
%%%%%%%%%%%%%%%%%%%%%%%%%%%%%%%%%%%%%%%%%%%%%%%%%%%%%%%%%%%%%%%%%%%%%%%%%%%%%%%
%%%%%%%%%%%%%%%%%%%%%%%%%%%%%%%%%%%%%%%%%%%%%%%%%%%%%%%%%%%%%%%%%%%%%%%%%%%%%%%
\newpage
\appendix
\onecolumn
\section{Appendix.}
\subsection{Additional Results}\label{app:results}
\subsubsection{Supervised fine-tuning.} To confirm that our \nlex data mix does not negatively impact the performance of the model on general tasks at the expense of boosting output prediction, we look at several standard coding and maths benchmarks. Specifically we consider \crux-Input \cite{cruxeval}, MBPP \cite{mbpp},  HumanEval Plus \cite{evalplus}, LiveCodeBench v5 \cite{lcb}, GSM8k \cite{gsm8k},  and Math 500 \cite{math500}. As reported in Table \ref{tab:sft_other_metrics}, using the \nlex data mix does not notably harm any metric, and even improves output prediction abilities as noted by \crux-Input.

\begin{table}[ht]
\centering
\caption{Investigating the impact of the \nlex data mix compared to a standard reasoning and instruction following mix on various standard coding and maths benchmarks. All models are trained with supervised fine tuning for the same budget only changing the data.}
\label{tab:sft_other_metrics}
\begin{tabular}{@{}lcccccc@{}}
\toprule
\textbf{Model} & \textbf{\crux-In} & \textbf{MBPP} & \textbf{HumanEval+} & \textbf{LCBv5} & \textbf{GSM8k} & \textbf{Math 500} \\ \midrule
Qwen2.5-7B regular mix   & 0.469 & 0.634 & 0.652 & 0.414 & 0.842 & 0.518 \\
\quad + \nlex & 0.505 & 0.632 & 0.659 & 0.413 & 0.826 & 0.528 \\
\midrule
Qwen2.5-3B regular mix   & 0.361 & 0.522 & 0.543 & 0.195 & 0.748 & 0.398 \\
\quad + \nlex & 0.445 & 0.524 & 0.537 & 0.203 & 0.729 & 0.406 \\ 
\bottomrule
\end{tabular}
\end{table}

\subsubsection{The effect of RL on output prediction}
\label{app:rl_abb}
To better evaluate the effect of additional RL phase on top of the SFT, we evaluate CWM model, trained with and without RL considering output prediction only. For a reference, we additionally include results for the official post-trained CWM model. Results are reported in~\cref{tab:out_pred_rl_abb}. As expected, results suggest that the RL phase significantly improve results on output prediction over competitive programming questions. We omit Qwen results as without RL, their performance was significantly lower.

\begin{table}[ht]
  \caption{Output prediction performance of the CWM with and without RL. For reference we additionally report results for the official CWM model.}
  \label{tab:out_pred_rl_abb}
  \begin{center}
    \begin{small}
      \begin{sc}
      \resizebox{0.6\columnwidth}{!}{
        \begin{tabular}{lcccccc}
          \toprule
          Base Model & \multicolumn{3}{c}{\lcbio-Out} & \multicolumn{3}{c}{DMC-Test-Out} \\
                & pass@1 & @5 & @10 & @1 & @5 & @10 \\
          \midrule
          CWM (Official)    & 60.4 & 79.3 & 82.0 & 68.9 & 85.8 & 87.8  \\
          \midrule
          CWM (wo. RL) & 30.3 & 55.6 & 62.1 & 38.4 & 67.5 & 73.5  \\
          CWM (w. RL) & \bf{89.6} & \bf{93.4} & \bf{94.2} & \bf{89.2} & \bf{93.3} & \bf{94.0}  \\
          \bottomrule
        \end{tabular}}
      \end{sc}
    \end{small}
  \end{center}
  \vskip -0.1in
\end{table}

\subsubsection{Self-Verification}
\begin{figure*}[t!]
  \begin{center}    \centerline{\includegraphics[width=\textwidth]{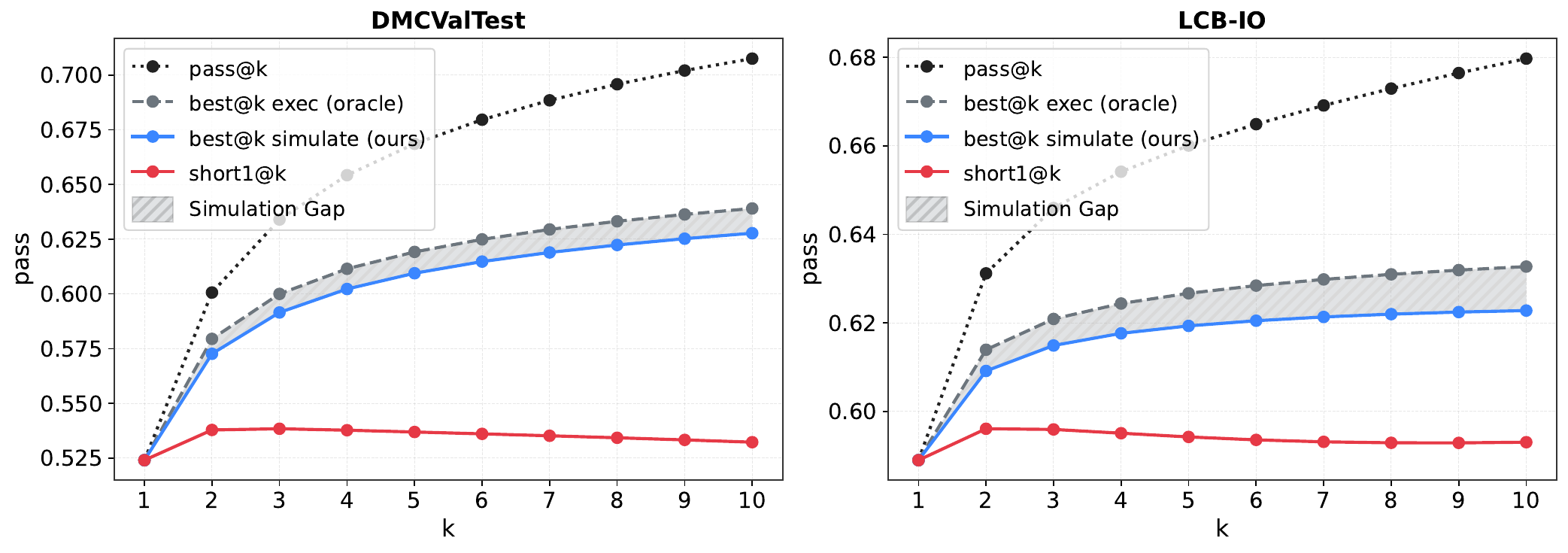}}
    \caption{Comparing best@k when ranking solutions generated by CWM post-trained jointly for solving and output prediction, using the same model as a verifier. The model here was trained and evaluated without the public tests as part of the description.}
    \label{fig:cwm_testless_self_verification}
  \end{center}
\end{figure*}

To further analyse the impact if self-verification using self-execution simulation, we wanted to study a setup where the tests used for verification were not present at the time of generating the solutions. To that end, we generate solutions using a model trained jointly for output prediction and competitive programming solving, but \textbf{without} tests in the question description for training and inference. This represents a case where the tests for verification contain completely new information unseen when generating the solution. Results are provided in Figure \ref{fig:cwm_testless_self_verification}. These suggest that while removing the tests from the description has a negative notable impact on performance, much of the performance can be gained by filtering solutions using these tests. It can also suggest that tests not used for generating the solution could have a higher positive impact for verification, motivating future investigation of test generation.

\subsubsection{Self-RLEF}\label{app:srlef}
By default we allow a maximum of $10$ turns for execution RLEF, and \emph{Self-RLEF}. However in practice the model often submits its solution prior leading to an average of $3.33$ turns. We wish to consider the performance of \emph{Self-RLEF} when limiting the number of solve turns to a maximum of $3$. We provide results in Table \ref{tab:ef_solving_fix2}. In practice the model uses an average of $2.38$ turns.

\begin{table*}[t!]
  \caption{Solve rates when using real or simulated execution feedback, but limiting to $3$ turns. This extends Table \ref{tab:ef_solving} under a more compute constraint setup.}
  \label{tab:ef_solving_fix2}
  \begin{center}
    \begin{small}
      \begin{sc}
        \begin{tabular}{l|cccccccc}
          \toprule
           & \multicolumn{4}{c}{\dmcvt}   & \multicolumn{4}{c}{\lcbio}  \\
          & pass@1 & pass@5 & pass@10 & public & pass@1 & pass@5 & pass@10 & public \\
          \midrule
          Self-RLEF (Ours) & 61.5 & 75.6 & 79.6 & 79.0 & 61.5 & 69.2 & 71.1 & 84.2 \\
          Execution RLEF (Oracle) & 62.7 & 75.8 & 78.8 & 81.5 & 63.3 & 70.3 & 72.2 & 86.3 \\
          \bottomrule
        \end{tabular}
      \end{sc}
    \end{small}
  \end{center}
  \vskip -0.1in
\end{table*}

\subsubsection{Beyond \emph{Self}-Verification}
\begin{figure*}[htpb]
  \begin{center}    \centerline{\includegraphics[width=0.8\textwidth]{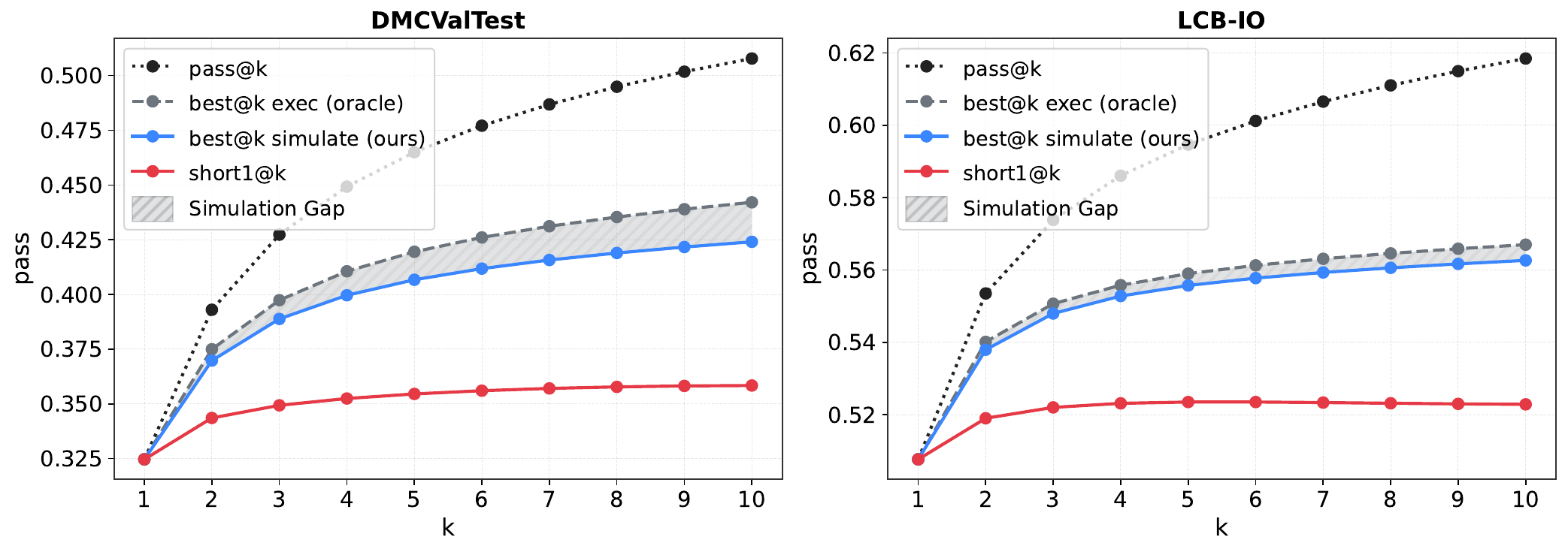}}
    \caption{Comparing best@k when ranking Qwen3-4B solutions, using CWM post-trained only for output prediction as a verifier. }
    \label{fig:qwen4_verification_cwm}
  \end{center}
\end{figure*}

\begin{figure*}[htpb]
  \begin{center}    \centerline{\includegraphics[width=0.8\textwidth]{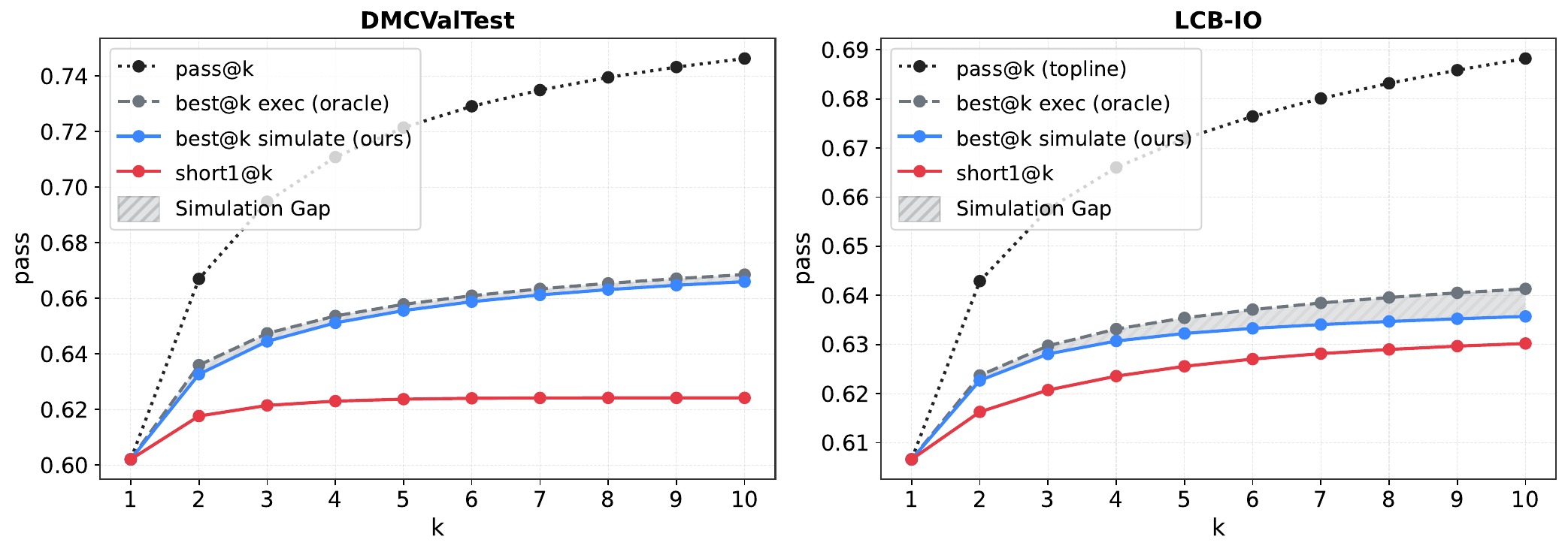}}
    \caption{Comparing best@k when ranking solutions by CWM post-trained only for competitive programming solving (denoted SOLVE-RL in Table \ref{tab:ef_solving}), using CWM post-trained only for output prediction as a verifier. }
    \label{fig:cwm_solve_verification_cwm}
  \end{center}
\end{figure*}

We provide results for using a dedicated output prediction model as a tool for verifying solutions of other models in a best@k setup. Results provided in Figures \ref{fig:qwen4_verification_cwm} and \ref{fig:cwm_solve_verification_cwm} show consistent improvements from this approach, for both Qwen3-4B and CWM Solve-RL, with only a slight degradation compared to ground truth execution of these tests. Like the results for Qwen3-32B in the main paper this further demonstrates the efficacy of this approach.

\begin{figure*}[t!]
  \begin{center}    \centerline{\includegraphics[width=0.8\textwidth]{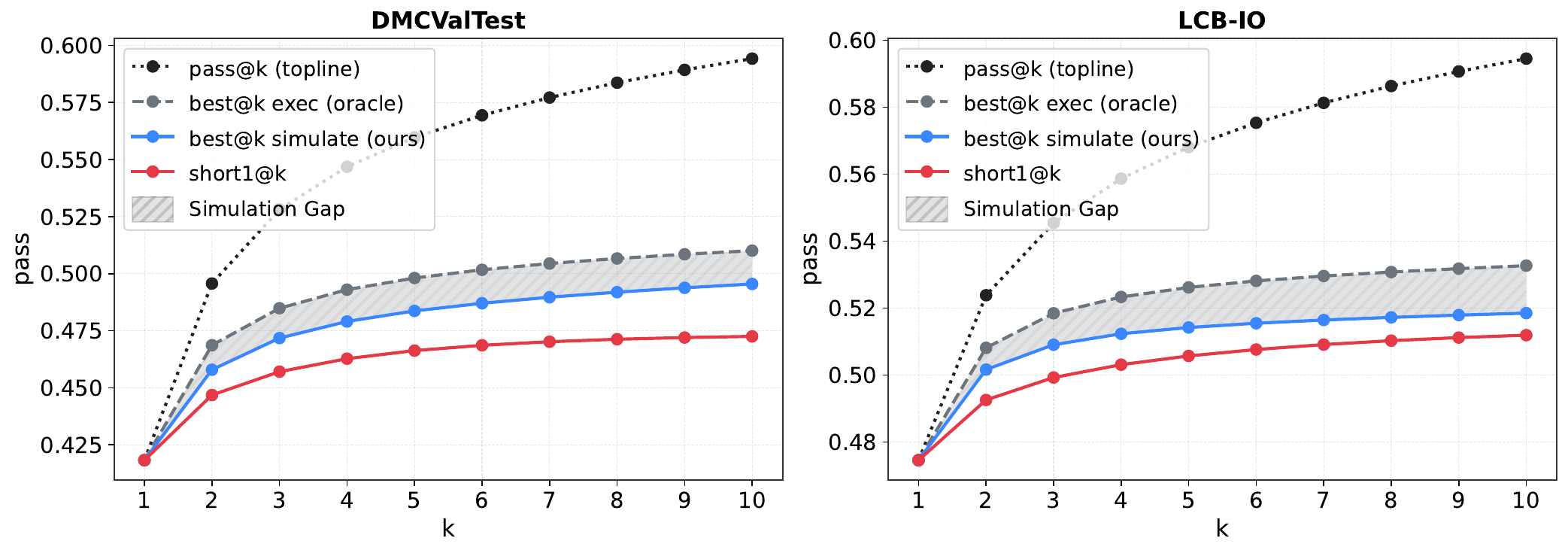}}
    \caption{Comparing best@k when ranking solutions by Qwen-7B post-trained for competitive programming solving, using Qwen-7B post-trained only for output prediction as a verifier. This mirrors the results for Qwen in Figure \ref{fig:self_verification}, but when each model has a dedicated role.}
    \label{fig:qwen_solve_verification_qwen_out}
  \end{center}
\end{figure*}

We also provide results of using a dedicated verifier based on a smaller model (Qwen2.5-7B), on solutions generated by a model starting from the same base model. Results provided in Figure \ref{fig:qwen_solve_verification_qwen_out} show that this method is also effective with models at this scale. This outperforms the performance in Figure \ref{fig:self_verification} which suggests that the constraint of having the same model for solving and verification does impose challenges especially with models with limited capacity.

\newpage
\subsection{Self-RLEF Example Inference}\label{app:srlef_example}
To demonstrate how the iterative self-fixing and self verification looks in practice in \emph{Self-RLEF} we provide the (abbreviated) multi-turn inference for a successful \lcbio solution using our model.

\begin{styletext}[title=Turn 1 - (Abbreviated) Solve Prompt]
Provide a Python solution for [...]: You are given two strings s and t, both consisting of lowercase English letters. You are going to type the string s character by character, from the first character to the last one.
When typing a character, instead of pressing the button corresponding to it, you can press the "Backspace" button. It deletes the last character you have typed among those that aren't deleted yet (or does nothing if there are no characters in the current string). For example, if s is "abcbd" and you press Backspace instead of typing the first and the fourth characters, you will get the string "bd" (the first press of Backspace deletes no character, and the second press deletes the character 'c'). Another example, if s is "abcaa" and you press Backspace instead of the last two letters, then the resulting text is "a". Your task is to determine whether you can obtain the string t, if you type the string s and press "Backspace" instead of typing several (maybe zero) characters of s.[...]

Output
For each test case, print "YES" if you can obtain the string t by typing the string s and replacing some characters with presses of "Backspace" button, or "NO" if not.

--------------- Example Tests------------
----- Test 0 -----
## Input
`4\nababa\nba\nababa\nbb\naaa\naaaa\naababa\nababa\n`

## Expected Output
`YES\nNO\nNO\nYES\n`.
\end{styletext}

\begin{stylecode}[title=Turn 1 - (Abbreviated) Solve Response]
[...Reasoning...]
import sys

def main():
    q = int(sys.stdin.readline())
    for _ in range(q):
        s = sys.stdin.readline().strip()
        t = sys.stdin.readline().strip()
        if len(t) > len(s):
            print("NO")
            continue
        current_states = {''}
        for c in s:
            new_states = set()
            for state in current_states:
                # Option 1: type the character
                new_state = state + c
                new_states.add(new_state)
                # Option 2: press backspace
                if len(state) > 0:
                    new_state_back = state[:-1]
                    new_states.add(new_state_back)
            current_states = new_states
        print("YES" if t in current_states else "NO")

if __name__ == "__main__":
    main()
\end{stylecode}

\begin{styletext}[title=Turn 2 - (Abbreviated) Simulation Response]
[...Simulation Reasoning...]
For each state in current_states (which is ['']), we have two options:
- Type 'a': new state is '' + 'a' = 'a'
- Press backspace: since state is empty, can't do this.
[...]
State '':
- Type 'a': '' + 'a' = 'a'
- Press backspace: can't, since state is empty.
[...]
Therefore, the final answer is:</think>
<output>NO\nNO\nNO\nNO\n</output>
\end{styletext}

\begin{styletext}[title=Turn 3 - (Abbreviated) Fix or Submit Prompt]
[...]
Review the provided code solution given the execution feedback and judge if it is correct or buggy. If the code is correct, end your response with a markdown comment EXACTLY ```#SUBMIT```. DO NOT repeat the given code. If the code is incorrect, provide a correct solution. [...]

Problem: [...]
Attempted Solution: [...]

----- Test 0 -----
## Input
`4\nababa\nba\nababa\nbb\naaa\naaaa\naababa\nababa\n`

## Expected Output
`YES\nNO\nNO\nYES\n`

## Output for the attempt
`NO\nNO\nNO\nNO\n`
\end{styletext}

\begin{styletext}[title=Turn 3 - (Abbreviated) Fix or Submit Prompt]
[...Reasoning...]
But according to the attempted solution, the output for this test case is NO, but the expected output is YES. So the attempted solution is not working correctly.
[...]
Wait, here's a problem! Ah! Here's the issue. In the attempted solution, when considering pressing Backspace, the code checks if len(state) > 0. If so, it adds state[:-1]. But when the state is empty, pressing Backspace does nothing, so the new state should still be empty. However, the code does not consider this case. [...]

```import sys

def main():
    q = int(sys.stdin.readline())
    for _ in range(q):
        s = sys.stdin.readline().strip()
        t = sys.stdin.readline().strip()
        if len(t) > len(s):
            print("NO")
            continue
        current_states = {''}
        for c in s:
            new_states = set()
            for state in current_states:
                # Option 1: type the character
                new_state = state + c
                new_states.add(new_state)
                # Option 2: press backspace
                new_state_back = state[:-1]
                new_states.add(new_state_back)
            current_states = new_states
        print("YES" if t in current_states else "NO")

if __name__ == "__main__":
    main()```
\end{styletext}

\begin{styletext}[title=Turn 4 - (Abbreviated) Simulation Response]
[...Simulation Reasoning...]
Therefore, the output is as I determined.
<output>YES\nNO\nNO\nYES\n</output>
\end{styletext}

\begin{styletext}[title=Turn 5 - (Abbreviated) Fix or Submit Response]
[...Reasoning...]
The given code seems to work for the provided test cases, as the output matches the expected output. [...]
Given that the code passes the provided test cases and the logic seems correct, I would judge the code as correct.</think>
```#SUBMIT```
\end{styletext}

\subsection{Hyper-Parameters}\label{app:hps}
\paragraph{Supervised Fine-Tuning.} All Qwen supervised fine-tuning use a sequence length of $65,536$ by applying scaled RoPE \cite{rope} with a factor of two relative to the base models to support longer context. CWM uses a context length of $131,072$ like in the original paper. Models were trained for $15.5$k steps, with a batch size of $4$M tokens per-update step, for a total of $65$B tokens. They were trained using a peak learning rate of $8e-6$ after a warmup of $1$k steps.
The estimated compute per-training run is $7.9e21$ FLOPs for 7B, and $5.0e21$ for 3B. Both models were trained for ${\sim}20$ hours on 128 and 64 NVIDIA H100 GPUs respectively. 

\paragraph{Reinforcement Learning.} We train the models on NVIDIA H100
GPUs, with a standard configuration of $192$ GPUs for a single training run of CWM, and $86$ for Qwen 7B and 3B. Typically 1/3 of the GPUs are used as trainers and the rest for rollouts. By default, we employ the maximum context of the model from SFT for generation, packing training sequences by  maximum of 131,072
tokens, use a global batch size of 1M tokens, a group size of 8, discarding rollouts with a staleness of more than 8 off-policy steps. We train the CWM models for 10k update steps, and the Qwen models for 4k, as we noted loops and collapses with longer training. This corresponds to roughly 9B and 3.2B tokens respectively. We use the last checkpoint for CWM, as training was stable, and the best checkpoint based on pass@1 by DMCValidation for Qwen (at 200 step increments) as the training was more prone to degradation in the end of training. We use 400 steps of linear learning rate warmup to a peak $1.4e-7$, with gradient clipping at $0.1$. 
For single turn solving jointly with output prediction we sample output prediction at $15\%$ of the time while the rest is for solving. For Self-RLEF we increase this ratio to $25\%$.

For sampling in evaluation we compare temperature $0.6$ and $1.0$, with top-p $0.95$ as these were common values for Qwen and CWM. We select the best per-model based on \dmcvt pass@1 rates. For CWM results didn't change notably for all training setups, and yet for Qwen with temperature of $0.6$ there were many loops leading to not finishing rollouts, this could be to the smaller model size. Thus for all Qwen models we use temperature $1.0$, as well as for all CWM models except of the results with two fixing turns which performed slightly better with $0.6$.

\subsection{Prompts}
\label{app:prompts}
As mentioned in Section \ref{sec:nlex} the data is created by converting raw traces to natural language by prompting an LLM, followed by a verification procedure. Below we provide the prompt used for the conversion.

\begin{styletext}[title=Trace to Natural Language Prompt]
You are an expert programmer tasked with explaining the step-by-step execution of a Python code snippet based on a provided execution trace. 
Focus and explain the specific values of variables at each step, not just vaguely say what the code does. Be specific about what the values of variables are. 
Note that the code could have bugs making it NOT do what the names suggest!
The trace shows <local> and <global> variables at each step, only where the values change, and the current <code> line.
Explain the provided execution. NEVER refer to or mention the trace itself. The output should ONLY be the execution explanation.
DO NOT use <> tags. You should adhere strictly to the trace for the execution, even if there is a mistake.
Do NOT suggest code fixes, even if the function is incorrect.

Finally, in addition to the explanation, provide the correct function output formatted as a valid Python literal, so that it can be easily verified using an assert statement (e.g. assert foo(inputs) == output). 
For example, if the output is a string, it should be enclosed in quotes; if it's a list it should be enclosed in square brackets and if a tuple in parentheses.

We have the following code:

```
{source_code}
```

And this line by line trace execution of running {func_name}({input_str}):
{stack_trace_string}

Explain the line by line execution of {func_name}({input_str}), followed by the correct output of the function. 

The output format should be: [EXPLANATION]EXECUTION_EXPLANATION[/EXPLANATION] [OUTPUT]EXECUTION_OUTPUT[/OUTPUT], where EXECUTION_OUTPUT is the result of running {func_name}({input_str}).
\end{styletext}

\subsection{Data Samples}\label{app:samples}

We provide examples from our \nlex data below.

\begin{stylecode}[title=Ex. 1 - Code + Input]
[PYTHON]
def maxSubArrayDP(arr):
    dp = [0] * len(arr)
    dp[0] = arr[0]
    result = arr[0]
    for i in range(1, len(arr)):
        dp[i] = max(arr[i], dp[i - 1] + arr[i])
        result = max(result, dp[i])
    return result
assert maxSubArrayDP([1, 0, 0, 0, 0, 0]) == ??
[/PYTHON]
\end{stylecode}

\begin{styletext}[title=Ex. 1 - Natural Language Execution Trace]
[THOUGHT]
The function is executed as follows:

1. The input array is `[1, 0, 0, 0, 0, 0]`. A variable `dp` is initialized with the same length as the array, filled with 0s.
2. The first element of `dp` is set to the first element of the input array, which is `1`.
3. A variable `result` is also initialized with the first element of the array, which is `1`, representing the current maximum subsequence sum.
4. The code enters a loop to process each element in the array starting from the second position.
5. At the first iteration (`i = 1`), the value of `dp[i]` is computed as the maximum of the current element `0` and the sum of the previous `dp` value `1` and the current element. So, `dp[1] = 1`. The `result` is updated to the maximum of itself and `dp[1]`, remaining `1`.
6. At the second iteration (`i = 2`), the value of `dp[2]` is computed similarly as the maximum of `0` and `1 (dp[i-1]) + 0`. So, `dp[2] = 1`. The `result` still stays at `1`.
7. This process is repeated for the remaining indices (3, 4, and 5).
   - For each, the value of `dp[i]` at `0` will be compared with the sum of the previous `dp[i-1]` value (which was updated in the prior step, incrementally adding 1) and `0`. Thus, `dp[i]` will become `1` for each index `i >= 2`.
   - The `result` continues to remain at `1` since all updated `dp[i]` values are equal to the current maximum.
8. After all iterations, the loop finishes, and the final value of `result` is returned, which is the maximum sum of any contiguous subarray of the input array.
[/THOUGHT]
[ANSWER]
assert maxSubArrayDP([1, 0, 0, 0, 0, 0]) == 1
[/ANSWER]
\end{styletext}

\begin{stylecode}[title=Ex. 2 - Code + Input]
[PYTHON]
def translate(value, to_min, to_max, from_min, from_max):
  '''
      Translate a value from one range to another
  '''
  # Figure out how 'wide' each range is
  to_span = to_max - to_min
  from_span = from_max - from_min

  spans_decimal = to_span / from_span

  # Convert the left range into a 0-1 range (float)
  scaled_value = int(round(float(value - to_min) * spans_decimal, 1) + 1)

  return scaled_value
assert translate(11, 0, 10, 0, 20) == ??
[/PYTHON]
\end{stylecode}

\begin{styletext}[title=Ex. 2 - Natural Language Execution Trace]
[THOUGHT]
The function `translate` is executed with the arguments 11, 0, 10, 0, 20.

1. At the start of the function, the local variables `value`, `to_min`, `to_max`, `from_min`, and `from_max` are assigned the values 11, 0, 10, 0, and 20 respectively.
2. On the sixth line of the function, the variable `to_span` is calculated by subtracting `to_min` from `to_max`, resulting in `10 - 0 = 10`.
3. On the seventh line, the variable `from_span` is calculated by subtracting `from_min` from `from_max`, resulting in `20 - 0 = 20`.
4. On the ninth line, the variable `spans_decimal` is calculated by dividing `to_span` by `from_span`, resulting in `10 / 20 = 0.5`.
5. On the twelfth line, a new variable `scaled_value` is calculated in several steps: 
   - First, the difference between `value` and `to_min` is calculated, resulting in `11 - 0 = 11`.
   - Then, this result is multiplied by `spans_decimal`, resulting in `11 * 0.5 = 5.5`.
   - The result is rounded to one decimal place, which may not be consistent with the code: `round(5.5, 1)` results in 5.5 again.  
   - One is then added to this result, giving `5.5 + 1 = 6.5`.
   - Finally, this is cast to an integer, `int(6.5) = 6`, since any decimal part is truncated.
6. On the fourteenth line, the `scaled_value` (6) is returned.
[/THOUGHT]
[ANSWER]
assert translate(11, 0, 10, 0, 20) == 6
[/ANSWER]
\end{styletext}

\begin{stylecode}[title=Ex. 3 - Code + Input]
[PYTHON]
def additionLossFunc(x, inc):
    y = []
    for i in x:
        y.append(inc*i*100)
    return y
assert additionLossFunc([1, 2, 3, 4, 5, 6, 7, 8, 9, 10], 1.25) == ??
[/PYTHON]
\end{stylecode}

\begin{styletext}[title=Ex. 3 - Natural Language Execution Trace]
[THOUGHT]
The function starts by initializing an empty list `y`. 

It then begins a loop using each element in the list `x` as the loop variable `i`. 

In the first iteration, the value of `i` is 1. The value 1.25 (which is `inc`) is multiplied by `i` and then by 100. The result is 125.0, which is appended to the list `y`.

In the second iteration, the value of `i` is 2. The calculation becomes 1.25 * 2 * 100, which gives 250.0. This is appended to the list `y`.

In the third iteration, `i` is 3. The value is calculated as 1.25 * 3 * 100, resulting in 375.0, and appended to `y`.

Continuing in the same way for the rest of the elements in the list, the following values are computed and appended to `y`:

- For `i` = 4, result is 500.0
- For `i` = 5, result is 625.0
- For `i` = 6, result is 750.0
- For `i` = 7, result is 875.0
- For `i` = 8, result is 1000.0
- For `i` = 9, result is 1125.0
- For `i` = 10, result is 1250.0

After all the iterations, the final list `y` contains all the computed results from each iteration. The function returns this list as the output.
[/THOUGHT]
[ANSWER]
assert additionLossFunc([1, 2, 3, 4, 5, 6, 7, 8, 9, 10], 1.25) == [125.0, 250.0, 375.0, 500.0, 625.0, 750.0, 875.0, 1000.0, 1125.0, 1250.0]
[/ANSWER]
\end{styletext}

%%%%%%%%%%%%%%%%%%%%%%%%%%%%%%%%%%%%%%%%%%%%%%%%%%%%%%%%%%%%%%%%%%%%%%%%%%%%%%%
%%%%%%%%%%%%%%%%%%%%%%%%%%%%%%%%%%%%%%%%%%%%%%%%%%%%%%%%%%%%%%%%%%%%%%%%%%%%%%%

\end{document}